\documentclass{article}

  \usepackage[main, final]{neurips_2026}

\usepackage[utf8]{inputenc} %
\usepackage[T1]{fontenc}    %
\usepackage{hyperref}       %
\usepackage{url}            %
\usepackage{booktabs}       %
\usepackage{amsfonts}       %
\usepackage{nicefrac}       %
\usepackage{microtype}      %
\usepackage{xcolor}         %
\usepackage{multirow}
\usepackage{array}
\usepackage{makecell}
\usepackage{graphicx}
\usepackage{amsmath}
\usepackage{enumitem}
\usepackage{bm}
\usepackage{url}
\usepackage{caption}
\usepackage{wrapfig}
\newif\ifshowrev
\showrevfalse
\ifshowrev
  \DeclareRobustCommand{\rev}[1]{{\color{red}#1}}
  \newcommand{\revon}{\color{red}}
  \newenvironment{revblock}{\color{red}}{}
\else
  \DeclareRobustCommand{\rev}[1]{#1}
  \newcommand{\revon}{}
  \newenvironment{revblock}{}{}
\fi
\newcommand{\crmarkbib}[1]{\expandafter\def\csname crrevisedbib@#1\endcsname{}}
\crmarkbib{selvaraju2017gradcam}
\AtBeginDocument{%
  \let\croriginalbibitem\bibitem
  \renewcommand{\bibitem}[2][]{%
    \ifcsname crrevisedbib@#2\endcsname\revon\else\normalcolor\fi
    \if\relax\detokenize{#1}\relax
      \croriginalbibitem{#2}%
    \else
      \croriginalbibitem[#1]{#2}%
    \fi
  }%
}
\newcommand{\etal}{\textit{et al}.}

\newcommand{\eg}{\textit{e}.\textit{g}.}
\title{Temporal Slice Learning for AI-Generated Video Detection with 400× Fewer FLOPs}

\author{%
  {Wei Shang} \qquad {Zhiyuan Li} \qquad {Shengfeng He\thanks{Corresponding author.}} \\
  {School of Computing and Information Systems} \\
  {Singapore Management University} \\
  {\texttt{weishang@smu.edu.sg, zhiyuanli@smu.edu.sg, shengfenghe@smu.edu.sg}}\\
  Project page: \url{https://shangwei5.github.io/TSC-Net/}.
}

\begin{document}

\maketitle

\begin{abstract}
\vspace{-2mm}Recent advances in generative video models have significantly improved visual realism, making detection increasingly dependent on subtle temporal inconsistencies arising from imperfect frame-to-frame coherence. However, these signals are weak and spatially localized, while most of the video content is redundant. This creates a fundamental mismatch: existing detectors rely on heavy backbones to model full videos, assuming increased capacity can capture such cues, which leads to high computational cost and limited scalability.
In this work, we rethink AI-generated video detection from a representation perspective. Instead of modeling entire videos, we propose to explicitly restructure them into compact temporal slice representations that isolate informative temporal evolution. Based on this idea, we introduce a lightweight framework using frequency-aware temporal slice consistency learning, where informative spatial rows and columns are aggregated over time to form slice-domain inputs. This reparameterization suppresses redundant appearance while preserving discriminative temporal dynamics, enabling efficient and targeted detection.
By constructing slices from both high- and low-frequency regions, our method exposes complementary temporal artifacts, including unstable details and abnormal motion smoothness. We instantiate this idea in a \textit{Temporal Slice Consistency Network}, which integrates task-driven slice localization, hierarchical positional encoding, and a lightweight Transformer to model cross-slice and cross-frequency dependencies with minimal overhead.
We further introduce \textit{AI-Artist}, a new benchmark with artist-curated videos from recent high-fidelity generators, including Seedance and Kling. Experiments on GenVideo and AI-Artist show that our method achieves strong performance while requiring over 400× fewer \rev{model-forward} FLOPs and 3.6× faster \rev{model} inference than the strong video-based baseline ReStraV \rev{(2.9× faster end to end, including preprocessing)}, enabling scalable and practical deployment.
\vspace{-2mm}
\end{abstract}

\begin{figure}[t]
	\centering
	\vspace{-5mm}\includegraphics[width=\linewidth]{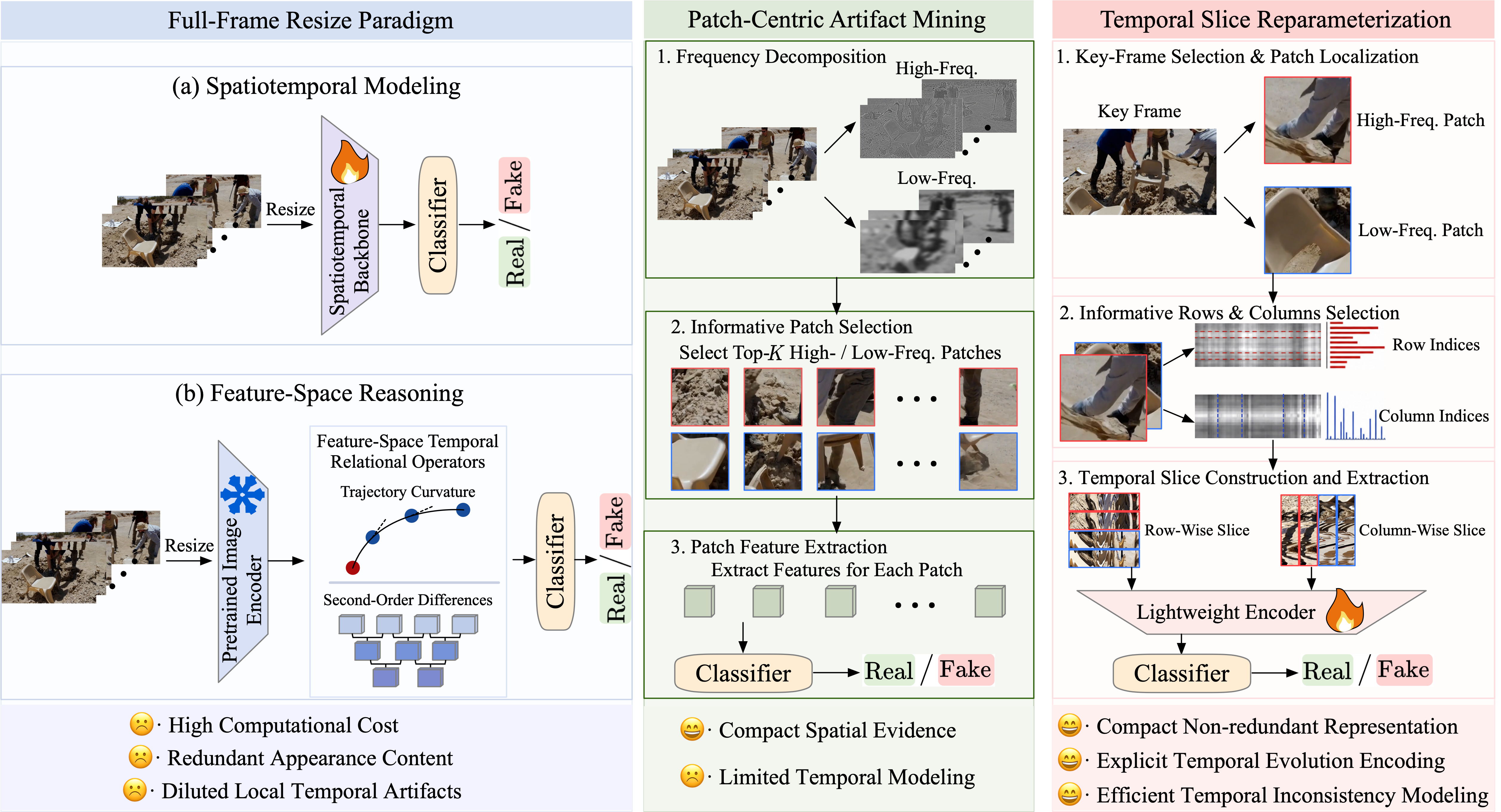} 
	\vspace{-6mm}\caption{Comparison of representative paradigms for AI-generated video detection. 
	Existing methods mainly rely on full-frame modeling or patch-level artifact mining. 
	In contrast, our method reformulates videos into compact frequency-aware temporal slices, where temporal evolution is explicitly encoded for lightweight and effective detection.}
	\label{fig:motivation}
	\vspace{-6mm}
\end{figure}

\section{Rethinking AI-Generated Video Detection}

Recent advances in generative video models have dramatically improved the realism of synthetic content, enabling highly convincing videos generated from text, images, or multimodal prompts. While these technologies create new opportunities for media creation, they also raise growing concerns regarding misinformation, media authenticity, and public trust. As a result, reliable AI-generated video detection has become an increasingly important problem~\cite{interno2025aigeneratedvideodetectionperceptual, zheng2025d3}.

Compared with AI-generated image detection, video detection introduces a distinct challenge: realism must be maintained not only spatially within each frame, but also temporally across consecutive frames. As frame-level artifacts become less obvious, the remaining traces of generation increasingly arise from imperfect temporal coherence, such as flickering textures~\cite{bu2025bytheway}, unstable structures~\cite{bai2025geovideo}, or overly smooth motion transitions~\cite{cuba2025flowmotion}. These cues are often weak, spatially localized, and difficult to perceive from isolated frames, making AI-generated video detection fundamentally different from image-level artifact detection.

A natural response is to scale up video modeling using spatiotemporal architectures~\cite{chen2024demamba} or large pretrained models~\cite{interno2025aigeneratedvideodetectionperceptual, zheng2025d3}. However, this strategy implicitly treats AI-generated video detection as a generic video understanding task, where richer semantic and motion representations are expected to improve recognition. We argue that this assumption is not fully aligned with the nature of the task. 
While global semantic and motion context can be useful, the most diagnostic cues for AI-generated video detection often lie in subtle generation inconsistencies hidden in local temporal evolution. Therefore, simply modeling full video volumes can introduce substantial appearance redundancy, dilute weak artifacts, and incur high computational cost.

\noindent\textbf{Observation 1: Detection benefits from emphasizing localized temporal inconsistency over full-video semantics.} %
As illustrated in Fig.~\ref{fig:motivation}, existing detectors are largely adapted from generic video recognition pipelines, where videos are processed as full-frame sequences~\cite{chen2024demamba, zheng2025d3, interno2025aigeneratedvideodetectionperceptual}. Such pipelines are designed to preserve rich appearance and motion information, but this is not always necessary for detecting generated videos. Since generation artifacts typically occupy only a small portion of the video volume, full-frame modeling spends most computation on normal scene appearance and high-level semantics. This redundancy increases FLOPs and latency, and may dilute the weak temporal signals that distinguish real and generated videos. Patch-centric artifact mining~\cite{yan2024sanity} offers more compact spatial evidence, but static patches alone may miss how local regions evolve over time.

\begin{figure}[!t]
	\centering
	\vspace{-5mm}\includegraphics[width=\linewidth]{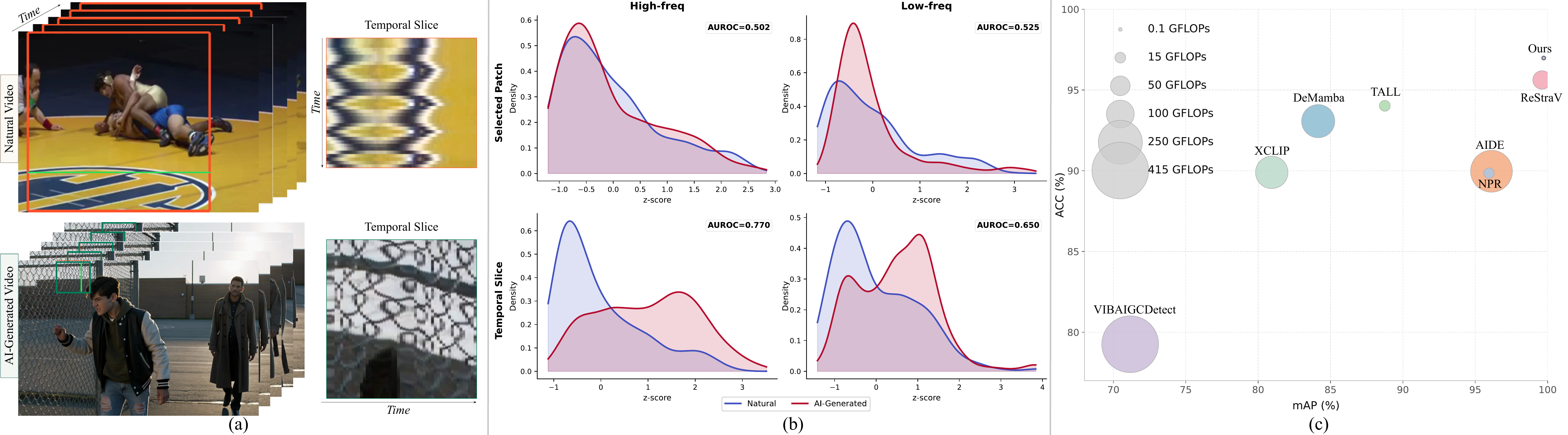} 
	\vspace{-6mm}\caption{
    Motivation and empirical evidence for temporal-slice-based AI-generated video detection.
(a) Temporal-slice visualization of natural and AI-generated videos. 
While natural videos tend to exhibit smoother and more coherent temporal evolution in static background regions, AI-generated videos may show temporal deformations and irregular textures that become more evident in slice form.
(b) Distributional analysis of natural (\textcolor{blue}{blue}) and AI-generated (\textcolor{red}{red}) videos in GenVideo~\cite{chen2024demamba} under patch-domain and slice-domain representations. 
Compared with selected spatial patches, temporal slices yield more distinguishable distributions, especially in high-frequency regions.
(c) Performance--efficiency comparison with representative detectors. 
The horizontal and vertical axes denote mAP and ACC \rev{on GenVideo}, respectively, while the bubble size indicates GFLOPs. 
Our method achieves a favorable trade-off, approaching the upper-right region with substantially lower computational cost.
    }
	\label{fig:intro_abc}
	\vspace{-7mm}
\end{figure}

\noindent\textbf{Observation 2: Slice-domain representations expose weak temporal cues compactly.}
The above limitations motivate us to rethink AI-generated video detection from a representation perspective. An effective representation should explicitly preserve local temporal evolution, suppress redundant scene appearance, and expose localized irregularities in a compact form. To this end, we reorganize informative spatial rows and columns across time into temporal slice representations. Unlike full-frame sequences, a temporal slice directly encodes how a selected local region changes over time, making irregular transitions easier to observe and model (as shown in Fig.~\ref{fig:intro_abc}(a)). This representation-level reparameterization is also computationally efficient, since temporal reasoning is performed on compact slice-domain inputs rather than dense video volumes.

Moreover, temporal artifacts may manifest differently across spatial frequency bands. High-frequency regions can reveal unstable details, boundary flickering, or texture-level jitter, while low-frequency regions may capture abnormal smoothness or coarse motion inconsistency. Therefore, we construct temporal slices from both high- and low-frequency regions, leading to a frequency-aware representation for temporal inconsistency modeling.

Fig.~\ref{fig:intro_abc}(b) supports this representation choice. In the patch domain, natural and AI-generated videos show heavy overlap in both high- and low-frequency regions, with AUROC scores close to random guessing ($0.502$ and $0.525$). After reorganizing the same content into temporal slices, separability improves substantially. High-frequency slices reach an AUROC of $0.770$, while low-frequency slices also improve to $0.650$. This analysis suggests that slice-domain representations expose temporal generation cues that are largely hidden in static patch statistics.

\noindent\textbf{Solution.}
Based on these observations, we formulate AI-generated video detection as \emph{frequency-aware temporal slice consistency learning}. Instead of processing entire video volumes, we construct compact slice-domain inputs from informative high- and low-frequency regions and directly model their consistency. \rev{A} prior face forgery study~\cite{gu2021spatiotemporal} used slice-like visualization to illustrate temporal inconsistency and further exploited feature-level temporal differences along horizontal and vertical directions. In contrast, our method treats temporal slices as the primary input representation for AI-generated video detection, rather than as an auxiliary visualization or an internal feature-difference operator.

Building on this formulation, we propose the \textit{Temporal Slice Consistency Network (TSC-Net)}, a lightweight framework that operates directly on frequency-aware temporal slices. TSC-Net integrates task-driven slice localization and assembly, hierarchical positional encoding, and a compact Transformer encoder to model cross-slice and cross-frequency dependencies. By performing temporal reasoning in the slice domain, TSC-Net avoids redundant full-video processing and achieves a favorable performance--efficiency trade-off. 
As shown in Fig.~\ref{fig:intro_abc}(c), our method lies within the top-performing region in terms of mAP and ACC, while requiring substantially fewer FLOPs than existing video-based detectors. 
This highlights the value of representation design: by preserving task-relevant temporal evidence while discarding redundant appearance content, compact slice-domain inputs enable competitive detection performance at a substantially lower computational cost.

In addition, existing benchmarks are increasingly saturated by earlier-generation video models and may not fully reflect the realism of recent generators. To better evaluate generalization \rev{on recent video generators}, we introduce \textit{AI-Artist}, a new benchmark containing artist-curated videos generated by recent high-fidelity models such as Kling 3.0~\cite{kling2026} and Seedance 2.0~\cite{seedance2026seedance}. 
Extensive experiments on both GenVideo and AI-Artist demonstrate that our method achieves strong overall performance while maintaining \rev{high} efficiency, requiring over 400× fewer \rev{model-forward} FLOPs than state-of-the-art video-based detectors such as ReStraV.

Our contributions are summarized as follows:
\begin{itemize}[leftmargin=*,itemsep=0pt,topsep=0pt]
\item We formulate AI-generated video detection as frequency-aware temporal slice consistency learning, where videos are reparameterized into compact slice-domain representations that explicitly preserve local temporal evolution.

\item We propose a lightweight Temporal Slice Consistency Network (TSC-Net), which integrates task-driven slice localization and assembly, hierarchical positional encoding, and cross-slice/cross-frequency dependency modeling.

\item We establish AI-Artist, a more challenging benchmark built from recent high-fidelity video generators, and provide extensive evaluations showing strong generalization and \rev{high} efficiency.
\end{itemize}

\section{Method}
\begin{figure}[!t]\footnotesize
	\centering
	\includegraphics[width=\linewidth]{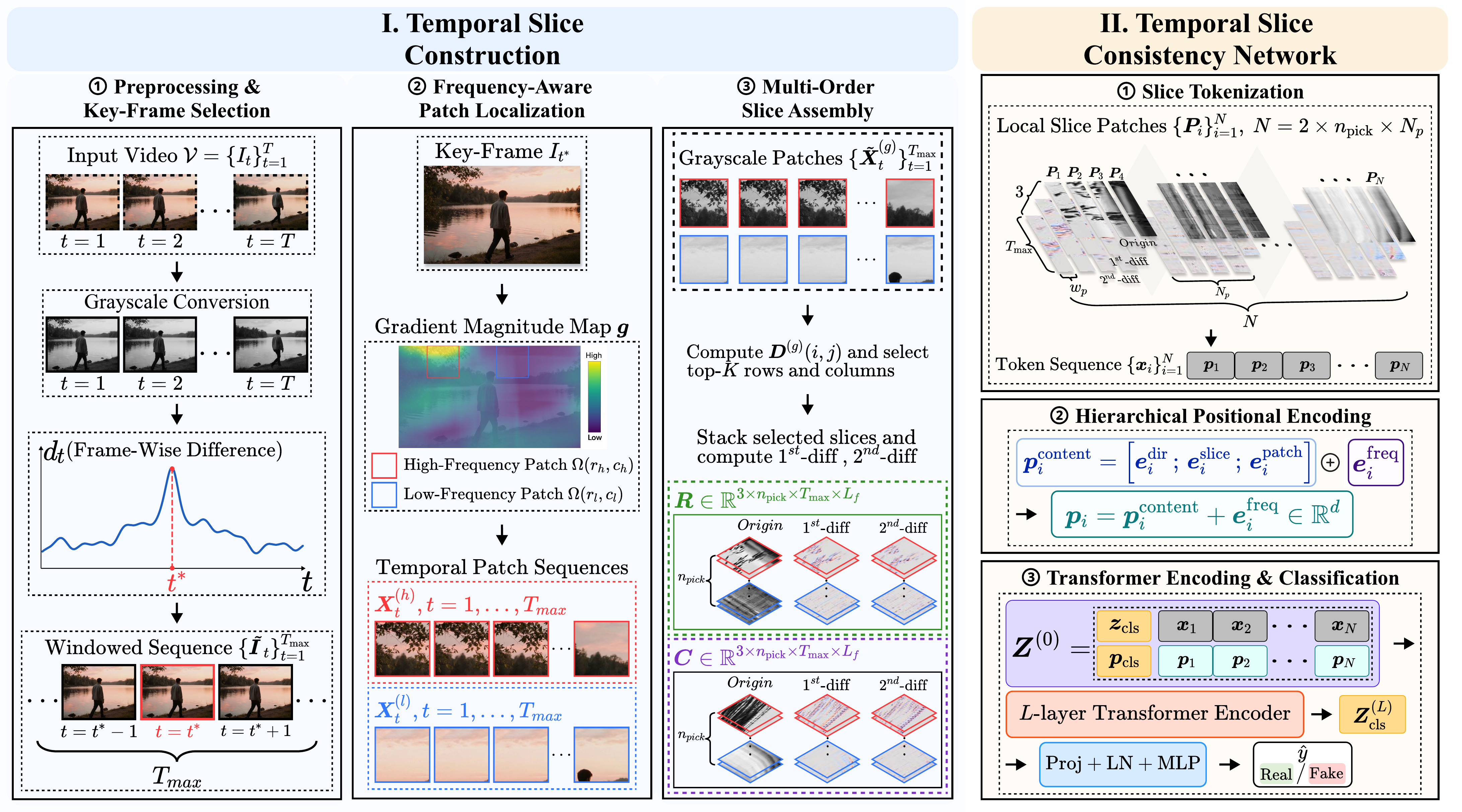} 
	\vspace{-6mm}\caption{Framework of our detection method TSC-Net.
	Given an input video, we first perform content-preserving preprocessing, key-frame selection, frequency-aware patch localization, and multi-order slice assembly to obtain compact temporal slice representations. 
	A lightweight Transformer then models cross-slice and cross-frequency dependencies for AI-generated video detection.}
	\label{fig:arch}
	\vspace{-\intextsep}
\end{figure}

We propose a lightweight AI-generated video detector based on \emph{frequency-aware temporal slices}, as shown in Fig.~\ref{fig:arch}. The core idea is to reparameterize an input video into a compact set of structured slice-domain representations and model their temporal consistency with a lightweight Transformer. 
Given an input video $\mathcal{V} = \{\bm I_t \}_{t=1}^{T}, \bm I_t \in \mathbb{R}^{3 \times H \times W}$,
our pipeline consists of two stages: 1) temporal slice construction, which includes content-preserving preprocessing, key-frame selection, frequency-aware patch localization, and slice assembly; 2) temporal slice consistency modeling with a lightweight Transformer.

\subsection{Temporal Slice Construction}
\label{sec:temporal_slice}

\subsubsection{Content-Preserving Preprocessing and Key-Frame Selection}
We first remove invalid borders according to the first frame and apply the same cropping window to all frames, preserving temporal alignment throughout the video.
To localize informative regions, we select a key frame based on maximal frame-wise temporal variation. Let $\bm G_t$ denote the grayscale version of frame $\bm I_t$.
We define the average inter-frame difference as
\begin{equation}
	d_t = \frac{1}{HW} \sum_{i=1}^{H}\sum_{j=1}^{W} \left| \bm G_{t+1}(i,j) - \bm G_t(i,j) \right|,
	\qquad t = 1, \dots, T-1.
		\label{eq:frame_diff}
\end{equation}
The key-frame index is then given by $t^\star = 1 + \arg\max_{t \in \{1,\dots,T-1\}} d_t$. Given $t^\star$, we extract a temporal window of length $T_{\max}$ centered at $t^\star$.
For videos longer than $T_{\max}$, frames outside the selected window are discarded; for shorter videos, padding is applied to maintain a fixed temporal length.
We denote the resulting windowed sequence by $\{\tilde{\bm I}_t\}_{t=1}^{T_{\max}}$.

\subsubsection{Frequency-Aware Patch Localization}
Temporal artifacts may appear differently across spatial frequency bands. 
High-frequency regions are often sensitive to unstable details and boundary flickering, whereas low-frequency regions can capture coarse motion inconsistency and abnormal smoothness. 
To incorporate these complementary cues, we localize one high-frequency patch and one low-frequency patch from the selected key frame.
Given the key frame ${\bm I}_{t^\star}$, we compute its gradient magnitude map $\bm g$. For each candidate $P \times P$ patch $\Omega(r,c)$, whose top-left corner is located at $(r,c)$, we compute its gradient energy as
	 $S(r,c) = \sum_{(i,j)\in \Omega(r,c)} \bm{g}(i,j)$.
The high-frequency and low-frequency patches are selected as
\begin{equation}
	(r_h, c_h) = \arg\max_{(r,c)}  S(r,c), \qquad
	(r_l, c_l) = \arg\min_{(r,c)}  S(r,c).
		\label{eq:patch_sel}
\end{equation}
Using these two fixed spatial locations, we extract two temporally aligned patch sequences from the windowed sequence:
\begin{equation}
	\bm X_t^{(h)} = \tilde{\bm I}_t[:,\, \Omega(r_h,c_h)], 
	\qquad
	\bm X_t^{(l)} = \tilde{\bm I}_t[:,\, \Omega(r_l,c_l)],
	\qquad t = 1,\dots,T_{\max},
\end{equation}
where $\bm X_t^{(h)}, \bm X_t^{(l)} \in \mathbb{R}^{3 \times P \times P}$.
This preserves spatial correspondence over time for subsequent slice construction.

\subsubsection{Multi-Order Slice Assembly}
We construct temporal slices separately from the high-frequency and low-frequency patch sequences.
Let $g \in \{h,l\}$ denote the frequency group, and let
$\{\bm X_t^{(g)}\}_{t=1}^{T_{\max}}$ be the corresponding RGB patch sequence. We first convert the patches to grayscale $\{\tilde{\bm X}_t^{(g)}\}_{t=1}^{T_{\max}}$.
To identify the most temporally informative rows and columns, we compute the mean absolute temporal difference map:
\begin{equation}
	\bm D^{(g)}(i,j)
	=
	\frac{1}{(T_{\max}-1)}
	\sum_{t=1}^{T_{\max}-1}
	\left|
	\tilde{\bm X}_{t+1}^{(g)}(i,j) - \tilde{\bm X}_t^{(g)}(i,j)
	\right|.
\end{equation}
Based on $\bm D^{(g)}$, we score each row and column by
\begin{equation}
	s_{\mathrm{row}}^{(g)}(i) = \frac{1}{P}\sum_{j=1}^{P} \bm D^{(g)}(i,j),
	\qquad
	s_{\mathrm{col}}^{(g)}(j) = \frac{1}{P}\sum_{i=1}^{P} \bm D^{(g)}(i,j).
\end{equation}
For each frequency group, we select the top-$K$ rows and the top-$K$ columns from $\{\tilde{\bm X}_t^{(g)}\}_{t=1}^{T_{\max}}$, where $K = n_{\text{pick}}/2$. We construct the temporal slice by stacking row-wise or column-wise pixel values across time.
After concatenating the two frequency groups, each of the row-wise and column-wise branches contains $n_{\text{pick}}$ slices in total.
Taking the row-wise branch as an example, stacking the selected row-wise slices yields $\bm S^{(0)} \in \mathbb{R}^{n_{\text{pick}} \times T_{\max} \times P}$.
We further compute first- and second-order temporal differences recursively:
\begin{equation}
	\bm S^{(k)}(:,t,:) =
	\begin{cases}
		\bm 0, & t=1,\\
		\bm S^{(k-1)}(:,t,:) - \bm S^{(k-1)}(:,t-1,:), & t>1,
	\end{cases}
	\qquad k \in \{1,2\}.
\end{equation}
To obtain a fixed tokenization length, all slices are aligned to a spatial length $L_f$ via cropping or padding.
We then stack the original slice, first-order temporal difference, and second-order temporal difference along the order dimension.
Thus, the row-wise temporal slice tensor is denoted as
$\bm R \in \mathbb{R}^{3 \times n_{\text{pick}} \times T_{\max} \times L_f}$.
Similarly, we obtain the column-wise temporal slice tensor
$\bm C \in \mathbb{R}^{3 \times n_{\text{pick}} \times T_{\max} \times L_f}$.

\subsection{Temporal Slice Consistency Network}

We model the resulting slice representation with a lightweight Transformer, termed the
\emph{Temporal Slice Consistency Network (TSC-Net)}.
The goal is to jointly reason over
(1) temporal consistency among slices from the same spatial direction,
(2) complementary information between row-wise and column-wise slices, and
(3) the distinction between high-frequency and low-frequency slice groups.

\subsubsection{Slice Tokenization}
Each row-wise or column-wise temporal slice in $\bm{R}$ and $\bm{C}$ is further divided along the spatial dimension into non-overlapping patches of width $w_p$, resulting in $N_p = \frac{L_f}{w_p}$ patches per slice. 
Collecting all local slice patches from $\bm R$ and $\bm C$ in a fixed order yields a sequence of $N$ local slice patches
$\{\bm P_i\}_{i=1}^{N}$, where $N = 2 \times n_{\text{pick}} \times N_p$.
Each local slice patch $\bm P_i \in \mathbb{R}^{3 \times T_{\max} \times w_p}$ is vectorized and projected into a $d$-dimensional token via a linear embedding:
\begin{equation}
	\bm x_i = \bm W_e \; \mathrm{vec}(\bm P_i), \qquad i = 1,\dots,N,
\end{equation}
where $\bm x_i \in \mathbb{R}^{d}$ and $\mathrm{vec}(\cdot)$ denotes vectorization.
This produces the token sequence $\{\bm x_i\}_{i=1}^{N}$.

\subsubsection{Frequency-Aware Hierarchical Positional Encoding}

Each token $\bm x_i$ is associated with multiple structural attributes of its source slice patch, including its spatial direction (row or column), slice index, local patch position within the slice, and frequency group (high-frequency or low-frequency). To preserve these cues, we construct a frequency-aware hierarchical positional encoding for each token.
\vspace{-\intextsep}
\paragraph{Frozen structural encoding.}
For the $i$-th token, we construct a deterministic structural encoding
$\bm p_i^{\mathrm{content}} =
	\left[
	\bm e_i^{\mathrm{dir}} \, ; \,
	\bm e_i^{\mathrm{slice}} \, ; \,
	\bm e_i^{\mathrm{patch}}
	\right]$.
Specifically, $\bm e_i^{\mathrm{dir}}$, $\bm e_i^{\mathrm{slice}}$, and $\bm e_i^{\mathrm{patch}}$ are fixed sinusoidal encodings of slice direction, slice index within its branch, and local patch location, with respective dimensions \rev{85, 85, and 86} ($d=256$).
\vspace{-\intextsep}
\paragraph{Learnable frequency embedding.}
To explicitly distinguish high-frequency and low-frequency slices, we introduce a learnable frequency embedding $\bm e_i^{\mathrm{freq}} \in \mathbb{R}^{d}$. Each token inherits the frequency label of its source slice. The final positional encoding for the $i$-th token is defined as
$\bm p_i = \bm p_i^{\mathrm{content}} + \bm e_i^{\mathrm{freq}}$, $i = 1,\dots,N$,
where $\bm p_i \in \mathbb{R}^{d}$.

Compared with fully learnable positional embeddings, this design disentangles deterministic structural coordinates from learnable frequency-group identity. 
The frozen structural encoding preserves the geometry of slice construction, while the learnable frequency embedding allows the model to adaptively calibrate the relative importance of high- and low-frequency cues. 
Empirically, this design improves detection performance and generalization, as validated in Sec.~\ref{app:ablation_comp}.

\subsubsection{Transformer Encoding and Classification}
We prepend a learnable classification token $\bm z_{\mathrm{cls}} \in \mathbb{R}^{d}$ to the token sequence.
A learnable positional embedding $\bm p_{\mathrm{cls}} \in \mathbb{R}^{d}$ is further assigned to this token.
The input to the Transformer encoder is
\begin{equation}
	\bm Z^{(0)} =
	\left[
	\bm z_{\mathrm{cls}};\,
	\bm x_1; \dots; \bm x_N
	\right]
	+
	\left[
	\bm p_{\mathrm{cls}};\,
	\bm p_1; \dots; \bm p_N
	\right], \quad
	\bm Z^{(0)} \in \mathbb{R}^{(N+1)\times d}.
\end{equation}
The sequence is then processed by an $L$-layer Transformer encoder, and the output state of the classification token is used for prediction:
\begin{equation}
	\bm Z^{(L)} = \mathrm{Transformer}\!\left(\bm Z^{(0)}\right), \qquad
	\hat{y} = \sigma\!\left(\mathrm{MLP}\!\left(\mathrm{Proj}\!\left(\mathrm{LN}\!\left(\bm Z^{(L)}_{\mathrm{cls}}\right)\right)\right)\right).
\end{equation}
The network is trained using the binary cross-entropy loss between $\hat{y}$ and the ground-truth label.

\section{Experiments}

\textbf{Training data.}
We follow the GenVideo training setting introduced in DeMamba~\cite{chen2024demamba} for a fair comparison. 
Specifically, the training data consists of real videos from Youku-mPLUG~\cite{xu2023youku} and fake videos generated by the same ten training generators used in GenVideo, including ZeroScope~\cite{zeroscope2024}, I2VGen-XL~\cite{wang2023i2vgenxl}, SVD~\cite{blattmann2023stable}, VideoCrafter~\cite{chen2024videocrafter2},
Pika~\cite{pika2022}, DynamiCrafter~\cite{xing2024dynamicrafter}, SD~\cite{zhang2024pia}, SEINE~\cite{chen2023seine}, Latte~\cite{ma2025latte}, and OpenSora~\cite{opensora2024}.
This training setup covers diverse generation paradigms while remaining consistent with the standard protocol adopted in prior work.

\textbf{Test data.}
We evaluate our method on two test benchmarks, both of which are strictly unseen during training.
\textbf{(1) GenVideo OOD test set.}
We first evaluate on the test set of GenVideo~\cite{chen2024demamba}. 
This set contains real videos from MSR-VTT~\cite{xu2016msr} and fake videos generated by ten unseen generators, including ModelScope~\cite{wang2023modelscope}, MorphStudio~\cite{morphstudio2023}, MoonValley~\cite{moonvalley2022}, HotShot~\cite{hotshot2023}, Show\_1~\cite{zhang2025show}, Gen2~\cite{esser2023structure}, Crafter~\cite{chen2023videocrafter1}, Lavie~\cite{wang2025lavie}, Sora~\cite{Brooks2024sora}, and WildScrape.
Importantly, this test split is explicitly out-of-domain with respect to training: both the real video source and the fake-video generators are disjoint from those used in the training set. 
Therefore, evaluation on GenVideo primarily measures cross-source and cross-generator generalization rather than in-distribution recognition.
\textbf{(2) AI-Artist Dataset.}
To evaluate detector generalization against the rapid evolution of generative models, we introduce \textit{AI-Artist}, a challenging benchmark of 380 videos generated by state-of-the-art models absent from GenVideo. 
To ensure a comprehensive evaluation scenario, the dataset is constructed via two complementary pipelines. 
The first is a \textit{web-collected} subset (221 videos) sourced from public communities (\eg, Discord, Douyin), designed to reflect the realistic in-the-wild distribution of AI-generated content (e.g., Kling 3.0~\cite{kling2026}) shared by online users. 
The second is a \textit{self-generated} subset (159 videos) systematically created via recent commercial models, such as Runway Gen-4.5~\cite{Runway2025}, PixVerse V5.6~\cite{PixVerse2026}, Hailuo 2.3~\cite{Hailuo2025}, Seedance 2.0~\cite{seedance2026seedance}, and Luma Ray3.14~\cite{Luma2026}. For this subset, \rev{we use} LLM-constructed prompts that span a rich variety of dynamic human actions, object categories, and complex camera motions \rev{to increase semantic diversity}. 
AI-Artist contains only AI-generated videos, as its goal is to evaluate detector generalization to recent high-fidelity generators. For binary evaluation, we construct a class-balanced test set by pairing the 380 AI-Artist generated videos with 380 real videos sampled from the real video split of the test set in GenVideo. The same sampled real set is used for all compared methods and all reported metrics. 
Detailed statistics and the full collection protocol are provided in Secs.~\ref{app:collection_data} and ~\ref{app:visualization_data}.

\textbf{Evaluation Metrics.}
Following prior work~\cite{chen2024demamba, interno2025aigeneratedvideodetectionperceptual, zhang2025towards, zheng2025d3}, we report \rev{Accuracy} (ACC), \rev{Average Precision} (AP), and the \rev{Area Under the Receiver Operating Characteristic Curve} (AUROC). AP and AUROC measure ranking performance without a fixed threshold; ACC uses a probability threshold of $0.5$ unless otherwise stated. \rev{Benchmark-level metrics are averaged across test subsets with weights proportional to their sample counts; the resulting mean AP and mean AUROC are denoted by mAP and mAUROC, respectively.} Model efficiency is evaluated in terms of the number of parameters, GFLOPs, and average inference time per video.
\begin{table}[!t]
	\caption{\rev{Comparison with state-of-the-art methods in terms of} \rev{Accuracy} (ACC), \rev{Average Precision} (AP), and Area Under the Receiver Operating Characteristic Curve (AUROC) on the GenVideo dataset. (I) and (V) denote image- and video-based methods, respectively. }
	\label{table:GenVideo}
	\centering
	\scriptsize
	\setlength{\tabcolsep}{3.8pt}
	\begin{tabular}{l l *{10}{c} c}
		\toprule
		\multirow{4}{*}{\textbf{Model}} &
		\multirow{4}{*}{\textbf{Metric}} &
		\multicolumn{10}{c}{\textbf{Dataset}} &
		\multirow{4}{*}{\textbf{Mean}} \\
		\cmidrule(lr){3-12}
		& & \textbf{Sora} & \makecell{\textbf{Morph}\\\textbf{Studio}} & \textbf{Gen2} & \textbf{HotShot} & \textbf{Lavie} & \textbf{Show-1} & \makecell{\textbf{Moon}\\\textbf{Valley}} & \textbf{Crafter} & \makecell{\textbf{Model}\\\textbf{Scope}} & \makecell{\textbf{Wild}\\\textbf{Scrape}} & \\
		\midrule
		\multirow{3}{*}{MLEP (I)}
		& ACC   & 0.1964 & 0.1214 & 0.1340 & 0.2586 & 0.1970 & 0.1429 & 0.0208 & 0.0770 & 0.3614 & 0.1384 & 0.1648 \\
		& AP    & 0.6180 & 0.6324 & 0.7211 & 0.6645 & 0.5559 & 0.6048 & 0.4855 & 0.5815 & 0.7319 & 0.6331 & 0.6229 \\
		& AUROC & 0.6116 & 0.6803 & 0.7943 & 0.6187 & 0.6098 & 0.6722 & 0.5580 & 0.6313 & 0.7564 & 0.6714 & 0.6604 \\
		\midrule
		\multirow{3}{*}{NPR (I)}
		& ACC   & 0.5536 & 0.9943 & 0.9970 & 0.7643 & 0.9710 & 0.9957 & 0.9984 & 0.9980 & 0.9829 & 0.7299 & 0.8985 \\
		& AP    & 0.8434 & 0.9971 & 0.9920 & 0.9122 & 0.9893 & 0.9898 & 0.9977 & 0.9985 & 0.9854 & 0.8892 & 0.9595 \\
		& AUROC & 0.8355 & 0.9963 & 0.9927 & 0.8984 & 0.9875 & 0.9912 & 0.9976 & 0.9982 & 0.9844 & 0.8328 & 0.9515 \\
		\midrule
		\multirow{3}{*}{VIBAIGCD (I)}
		& ACC   & 0.6786 & 0.8214 & 0.9040 & 0.7729 & 0.6980 & 0.8086 & 0.8514 & 0.8500 & 0.8171 & 0.7243 & 0.7926 \\
		& AP    & 0.6158 & 0.7220 & 0.8091 & 0.6679 & 0.6435 & 0.7028 & 0.7860 & 0.7678 & 0.7180 & 0.6826 & 0.7116 \\
		& AUROC & 0.6751 & 0.7456 & 0.8312 & 0.7112 & 0.6596 & 0.7344 & 0.7978 & 0.7790 & 0.7463 & 0.6776 & 0.7358 \\
		\midrule
		\multirow{3}{*}{AIDE (I)}
		& ACC   & 0.7679 & 0.9971 & 0.9990 & 0.9314 & 0.6090 & 0.9729 & \textbf{1.0000} & \textbf{1.0000} & 0.8514 & 0.8683 & 0.8997 \\
		& AP    & 0.8940 & 0.9979 & 0.9995 & 0.9776 & 0.8625 & 0.9836 & \textbf{0.9998} & 0.9993 & 0.9382 & 0.9592 & 0.9612 \\
		& AUROC & 0.8999 & 0.9979 & 0.9995 & 0.9784 & 0.8623 & 0.9851 & \textbf{0.9998} & 0.9993 & 0.9446 & 0.9501 & 0.9617 \\
		
		\midrule
		\multirow{3}{*}{TALL (V)}
		& ACC   & \textbf{1.0000} & 0.9971 & 0.9970 & 0.8614 & 0.9540 & 0.9414 & \textbf{1.0000}  & 0.9970 & 0.9914 & 0.8627 & 0.9602 \\
		& AP    & 0.8949 & 0.9763 & 0.9811 & 0.6846 & 0.8255 & 0.7969 & 0.9850 & 0.9723 & 0.9351 & 0.8232 & 0.8875 \\
		& AUROC & 0.8827 & 0.9729 & 0.9799 & 0.6913 & 0.8255  & 0.7888 & 0.9836 & 0.9697 & 0.9300 & 0.7733  & 0.8798\\
		\midrule
		
		\multirow{3}{*}{XCLIP (V)}
		& ACC   & 0.5714 & \textbf{1.0000} & \textbf{1.0000} & 0.5957 & 0.9840 & \textbf{1.0000} & 0.9936 & \textbf{1.0000} & 0.9986 & 0.8471 & 0.8990 \\
		& AP    & 0.4905 & 0.9904 & 0.9483 & 0.4141 & 0.9065 & 0.9662 & 0.7021 & 0.9801 & 0.9442 & 0.7515 & 0.8094 \\
		& AUROC & 0.4455 & 0.9899 & 0.9578 & 0.4075 & 0.9046 & 0.9665 & 0.7873 & 0.9803 & 0.9480 & 0.7108 & 0.8098  \\
		\midrule
		\multirow{3}{*}{DeMamba (V)}
		& ACC   & 0.6071 & \textbf{1.0000} & \textbf{1.0000} & 0.8600 & \textbf{0.9960} & \textbf{1.0000} & 0.9936 & \textbf{1.0000} & \textbf{1.0000} & \textbf{0.9498} & 0.9407 \\
		& AP    & 0.4722 & 0.9952 & 0.9875 & 0.5118 & 0.9554 & 0.9725 & 0.7541 & 0.9905 & 0.9495 & 0.8259 & 0.8415 \\
		& AUROC & 0.3887 & 0.9949 & 0.9892 & 0.5637 & 0.9558 & 0.9752 & 0.8097 & 0.9912 & 0.9498 & 0.7992 & 0.8418 \\
		\midrule
		\multirow{3}{*}{D3 (V)}
		& ACC   & --- & --- & --- & --- & --- & --- & --- & --- & --- & --- & --- \\
		& AP    & 0.9727 & 0.9762 & 0.9908 & 0.9772 & 0.9771 & 0.9827  & 0.9914  & 0.9844   & 0.9647   & 0.9232  & 0.9740  \\
		& AUROC & 0.9745 & 0.9714 & 0.9898 & 0.9721 & 0.9768 &  0.9804  &  0.9898  & 0.9817   & 0.9613  & 0.9246 & 0.9722  \\
		\midrule
		\multirow{3}{*}{ReStraV (V)}
		& ACC   & 0.9864 & 0.9835 & 0.9744 & 0.9864  & 0.9654 & 0.9864 & 0.9469 & 0.9844 & 0.9478 & 0.8101 & 0.9562 \\
		& AP    & 0.9977 & 0.9977 & 0.9976 & 0.9977 & 0.9773 & 0.9977 & 0.9994 & 0.9977 & 0.9969 & \textbf{0.9832} & 0.9961 \\
		& AUROC & 0.9977 & 0.9977 & 0.9976 & 0.9977 & 0.9973 & 0.9977 & 0.9995 & 0.9977 & \textbf{0.9969} & \textbf{0.9814} & 0.9959 \\
		\midrule
		
		\multirow{3}{*}{\textbf{Ours} (V)}
		& ACC   & \textbf{1.0000} & 0.9886 & 0.9890 & \textbf{0.9929} & 0.9640 & 0.9800 & 0.9920 & 0.9910 & 0.9657 & 0.8348 & \textbf{0.9698} \\
		& AP    & \textbf{1.0000} & \textbf{0.9996} & \textbf{0.9998} & \textbf{0.9999} & \textbf{0.9988} & \textbf{0.9994} & \textbf{0.9998} & \textbf{0.9997} & \textbf{0.9974} & 0.9782 & \textbf{0.9973} \\
		& AUROC & \textbf{1.0000} & \textbf{0.9996} & \textbf{0.9998} & \textbf{0.9999} & \textbf{0.9988} & \textbf{0.9994} & \textbf{0.9998} & \textbf{0.9996} & \textbf{0.9969} & 0.9728 & \textbf{0.9967} \\
		\bottomrule
	\end{tabular}
\vspace{-\intextsep}
\end{table}

\textbf{Implementation Details.}
All frames are processed using the preprocessing pipeline described in Sec.~\ref{sec:temporal_slice}. For each input video, we retain a temporal window of $T_{\max}=8$ frames centered at the selected key frame. In the temporal slice module, we set the number of selected slices to $n_{\text{pick}}=8$ and align each slice to a fixed spatial length of $L_f=64$. 
Each slice is further divided into non-overlapping local patches of width $w_p=8$ for tokenization. 
The Transformer encoder in TSC-Net uses an embedding dimension of $d=256$, with $4$ attention heads and $L=3$ encoder layers.
The batch size is set \rev{to} 32 \rev{during training}. The initial learning rate is set \rev{to} $3\times 10^{-4}$ with \rev{a cosine annealing schedule}~\cite{loshchilov2016sgdr}. \rev{We use} binary cross-entropy loss \rev{with the} AdamW optimizer~\cite{loshchilov2017decoupled} and \rev{train} all models \rev{for} 100K iterations. For competing methods, we follow their official implementations. All experiments are conducted on an NVIDIA RTX PRO 6000 GPU using PyTorch.

\subsection{Comparison with State-of-the-Art Methods on GenVideo}
We compare our method with representative image-level and video-level baselines, including MLEP~\cite{yuan2025mlep}, NPR~\cite{tan2024rethinking}, AIDE~\cite{yan2024sanity}, TALL~\cite{xu2023tall}, VIBAIGCD~\cite{zhang2025towards}, XCLIP~\cite{ni2022expanding}, DeMamba~\cite{chen2024demamba}, D3~\cite{zheng2025d3}, and ReStraV~\cite{interno2025aigeneratedvideodetectionperceptual}. The quantitative results on GenVideo are summarized in Table~\ref{table:GenVideo}, including ACC, AP, and AUROC on 10 datasets and their mean values. Note that D3 is a training-free ranking-based method that outputs the variance of second-order feature differences rather than calibrated classification probabilities. Therefore, threshold-dependent ACC is not directly applicable under its setting, and we follow the official protocol by reporting only AP and AUROC for D3.

\begin{table}[t]
	\centering
	\footnotesize
	\setlength{\tabcolsep}{4.0pt}
	\caption{Complexity and inference efficiency. \rev{GFLOPs and inference time are measured per video on GenVideo using the same measurement protocol for all methods: ``Prep.'' denotes method-specific, non-learnable data preprocessing, ``Model'' the model forward pass, and ``Total'' their sum. The last two columns report false positives on 10,000 real MSR-VTT clips at a threshold of $0.5$ (generated videos as the positive class); D3 is not included in this comparison because it outputs an uncalibrated ranking score.}}
	\label{table:complexity}
	\begin{tabular}{l c ccc ccc cc}
		\toprule
		\multirow{2}{*}{\textbf{Model}} & \multirow{2}{*}{\textbf{Params}} & \multicolumn{3}{c}{\textbf{GFLOPs}} & \multicolumn{3}{c}{\textbf{Inference Time (ms)}} & \multicolumn{2}{c}{\rev{\textbf{Real-video FP}}} \\
		\cmidrule(lr){3-5}\cmidrule(lr){6-8}\cmidrule(lr){9-10}
		& & \rev{Prep.} & \rev{Model} & \rev{Total} & \rev{Prep.} & \rev{Model} & \rev{Total} & \rev{\#FP} & \rev{FPR (\%)} \\
		\midrule
			MLEP (I) & 1.441\,M & \rev{0.012} & 14.35 & \rev{14.362} & \rev{7.975} & 5.998 & \rev{13.973} & \rev{422} & \rev{4.22} \\
			NPR (I) & \textbf{1.438}\,M & \rev{0.012} & 14.01 & \rev{14.022} & \rev{5.495} & 2.185 & \rev{7.680} & \rev{1,387} & \rev{13.87} \\
			VIBAIGCD (I) & 430.0\,M & \rev{0.012} & 415.2 & \rev{415.212} & \rev{8.322} & 30.50 & \rev{38.822} & \rev{4,813} & \rev{48.13} \\
			AIDE (I) & 897.8\,M & \rev{1.131} & 225.7 & \rev{226.831} & \rev{68.972} & 20.58 & \rev{89.552} & \rev{972} & \rev{9.72} \\
			\midrule
			TALL (V) & 86.92\,M & \rev{0.012} & 15.21 & \rev{15.222} & \rev{4.903} & 9.496 & \rev{14.399} & \rev{6,949} & \rev{69.49} \\
			XCLIP (V) & 121.3\,M & \rev{0.012} & 135.4 & \rev{135.412} & \rev{4.736} & 10.78 & \rev{15.516} & \rev{6,506} & \rev{65.06} \\
			DeMamba (V) & 125.7\,M & \rev{0.012} & 141.6 & \rev{141.612} & \rev{8.608} & 15.94 & \rev{24.548} & \rev{7,391} & \rev{73.91} \\
			D3 (V) & 121.3\,M & \rev{0.012} & 135.4 & \rev{135.412} & \rev{5.872} & 10.95 & \rev{16.822} & \rev{---} & \rev{---} \\
			ReStraV (V) & 20.06\,M & \rev{0.012} & 44.20 & \rev{44.212} & \rev{8.137} & 5.307 & \rev{13.444} & \rev{39} & \rev{0.39} \\
			\textbf{Ours} (V) & 1.731\,M & \rev{0.078} & \textbf{0.109} & \rev{\textbf{0.187}} & \rev{3.198} & \textbf{1.478} & \rev{\textbf{4.676}} & \rev{\textbf{7}} & \rev{\textbf{0.07}} \\
		\bottomrule
	\end{tabular}
	\vspace{-\intextsep}
\end{table}

As shown in Table~\ref{table:GenVideo}, our method achieves the best overall performance across all three metrics in terms of the mean over the 10 datasets. Specifically, it obtains a mean ACC of \textbf{0.9698}, outperforming ReStraV (0.9562) and TALL (0.9602) by \textbf{1.36} and \textbf{0.96} \rev{percentage points}, respectively. For AP, our method reaches \textbf{0.9973}, improving over the previous best ReStraV (0.9961) by \textbf{0.12} \rev{percentage points}, and surpassing D3 (0.9740) by \textbf{2.33} \rev{percentage points}. Similarly, in terms of AUROC, our method achieves \textbf{0.9967}, exceeding ReStraV (0.9959) by \textbf{0.08} \rev{percentage points} and D3 (0.9722) by \textbf{2.45} \rev{percentage points}. Moreover, beyond the averaged results, our method consistently ranks among the top-performing approaches on nearly all individual datasets, indicating strong generalization across diverse generation sources.
Additional feature-space visualizations are provided in Sec.~\ref{app:feature_visualization}, where the observed trends are consistent with the quantitative improvements in Table~\ref{table:GenVideo}.

We also evaluate computational efficiency in Table~\ref{table:complexity}. 
\rev{For all methods, the model forward pass and the method-specific preprocessing are measured separately; the preprocessing involves no learnable parameters.}
The efficiency advantage of our method is consistent across different families of detectors. 
Compared with the strong video-based detector ReStraV, our method achieves better overall detection performance while requiring \textbf{11.6× fewer} parameters, over \textbf{400× fewer} \rev{model-forward} FLOPs \rev{(0.109 vs.\ 44.20 GFLOPs)}, and \textbf{3.6× faster} \rev{model} inference. 
\rev{When preprocessing is included, our full pipeline still requires \textbf{236× fewer} GFLOPs and runs \textbf{2.9× faster} end to end (4.676 vs.\ 13.444\,ms), the lowest total cost among all compared methods. The speedup is smaller than the FLOPs reduction because fixed costs such as memory access and kernel launches account for a larger share of latency at low computational loads; a stage-wise breakdown is given in Sec.~\ref{app:pipeline_cost}.}

Compared with generic video or feature-space reasoning models such as XCLIP, DeMamba, and D3, the \rev{model-forward} FLOPs reduction further exceeds \textbf{1,200×}. 
Our method is also substantially more efficient than large image-based detectors such as AIDE and VIBAIGCD, which rely on heavy pretrained representations or high-capacity backbones. 
These comparisons indicate that TSC-Net obtains a favorable accuracy--efficiency trade-off not by increasing model scale, but by using a compact slice-domain representation that preserves task-relevant temporal evidence while suppressing redundant full-frame appearance.

\begin{wrapfigure}{R}{0.7\textwidth} 
	\vspace{-6mm}
		\centering
		\includegraphics[width=\linewidth]{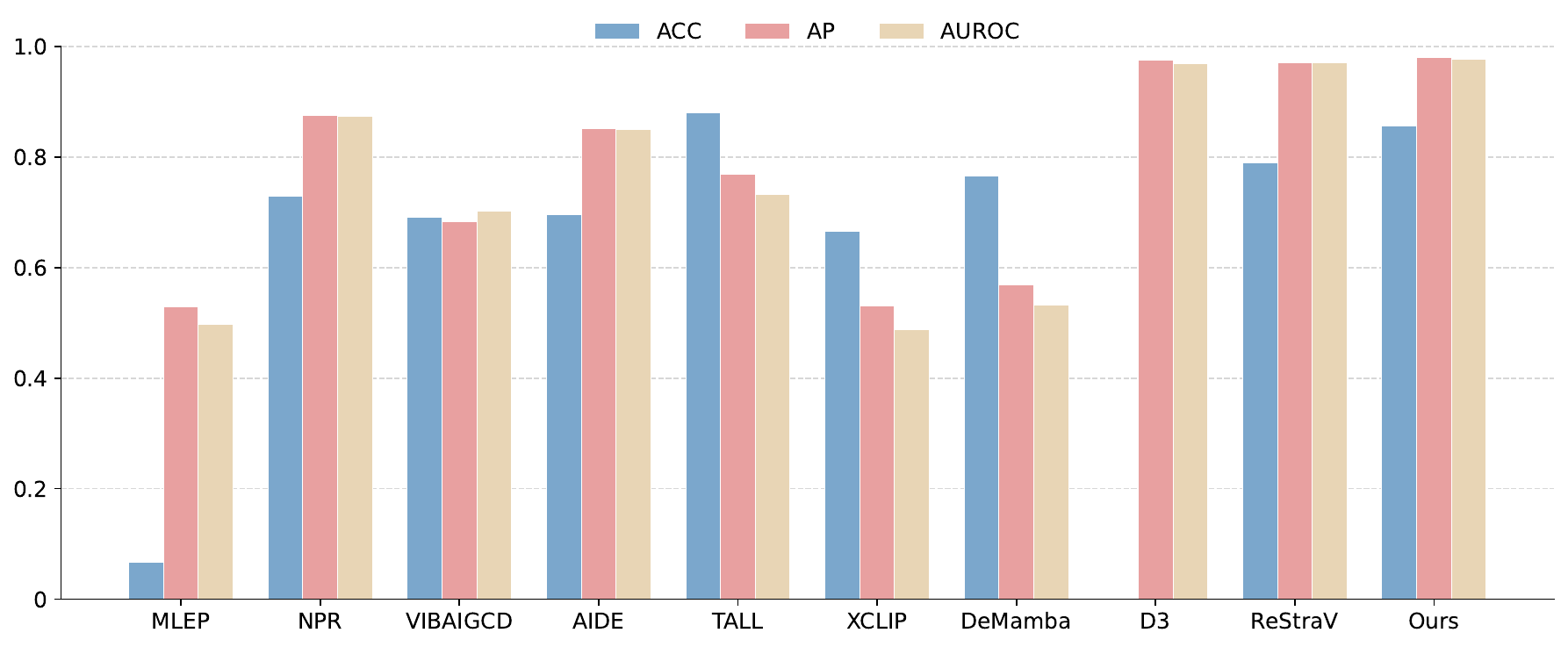} 
        \vspace{-6mm}
		\caption{The comparison of various models on our AI-Artist Dataset.}
		\label{fig:sota_extra}  \vspace{-3mm}
\end{wrapfigure}

\subsection{Evaluation on Our AI-Artist Dataset}

Fig.~\ref{fig:sota_extra} presents the comparison of different methods on our AI-Artist dataset, which contains videos generated by recent state-of-the-art models. Our method achieves the best overall performance, obtaining the highest AP and AUROC while maintaining competitive ACC. Although TALL yields the highest ACC, its AP and AUROC are markedly lower than those of our method, suggesting that its seemingly strong accuracy does not translate into equally reliable ranking and discrimination capability. In contrast, our method attains an ACC of \textbf{0.8558} and reaches nearly \textbf{0.98} on both AP and AUROC, surpassing strong baselines such as D3 and ReStraV (around 0.97) by about \textbf{1} \rev{percentage point}. These results further demonstrate the effectiveness and generalization ability of our method on more challenging and up-to-date generation distributions. Detailed per-generator results are provided in the supplementary material.

\subsection{Robustness to Post-processing Operations}

\begin{figure}[t]\footnotesize
	\centering
	\includegraphics[width=\linewidth]{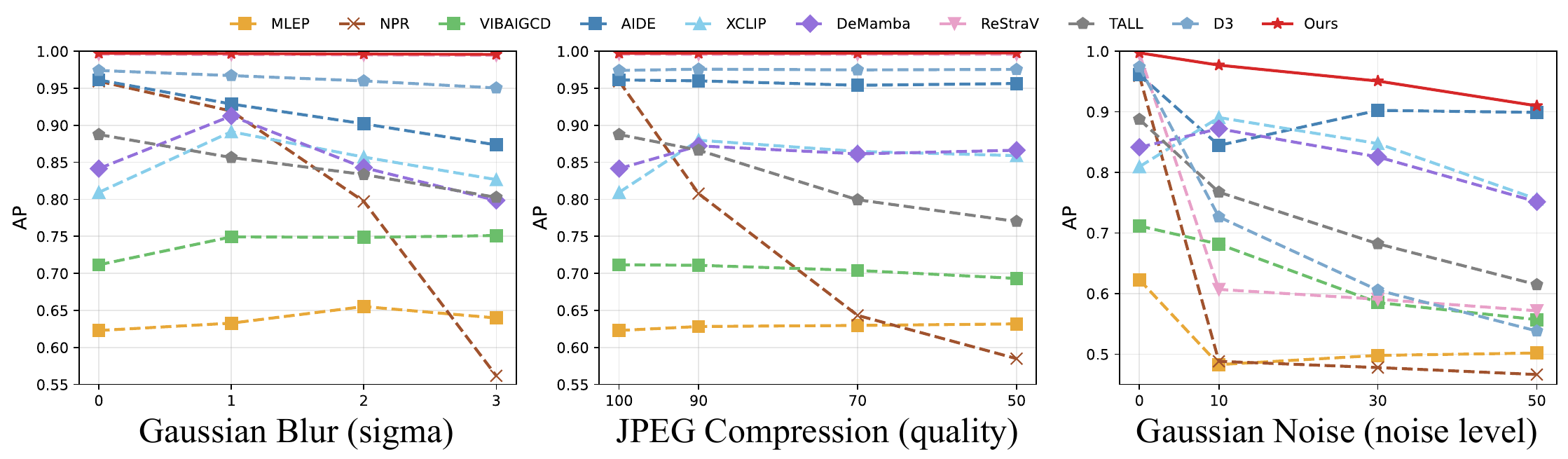} 
	\vspace{-6mm}
    \caption{Detection results of different detectors against post-processing operations on GenVideo, including Gaussian blur, JPEG compression, and Gaussian noise.}
	\label{fig:robust}
	\vspace{-\intextsep}
\end{figure}

In real-world scenarios, videos are rarely pristine and are often affected by post-processing operations, which may degrade detector reliability. To better assess robustness under such perturbations, we adopt AP rather than ACC as the evaluation metric. We observe that some methods tend to classify nearly all degraded samples as AI-generated; although it can yield high ACC, it reflects poor discriminative ability and usually corresponds to low AP. Therefore, AP provides a more faithful measure of robustness in this setting.
Fig.~\ref{fig:robust} reports the AP of different methods under Gaussian blur, JPEG compression, and Gaussian noise. Overall, our method consistently achieves the highest AP and exhibits the strongest stability across different degradation levels. Under JPEG compression, most methods remain relatively robust, while our method stays almost unchanged across all quality factors. Under Gaussian blur, several baselines show noticeable fluctuations or performance degradation, whereas our method remains highly stable. Gaussian noise is the most challenging perturbation for all detectors; nevertheless, our method still preserves comparatively stable performance and maintains the best AP throughout. These results indicate that our method not only performs strongly on clean videos, but also generalizes more reliably under post-processing distortions.

\begin{revblock}
\subsection{False Positives on Real Videos}
\label{sec:fpr_main}
We evaluate false positives on 10,000 real MSR-VTT clips (7,180 source videos, 20 categories), retaining each method's original score direction and a fixed threshold of $0.5$, without test-set tuning.
TSC-Net produces only 7 false positives (FPR $0.07\%$, specificity $99.93\%$), compared with 39 for ReStraV ($0.39\%$) and 422--7,391 for the other detectors (Table~\ref{table:complexity}).
We form six overlapping 2,500-clip subsets using the top or bottom quartile of fixed statistics on the 8-frame input window: strong temporal change, high contrast, rich high-frequency content, camera motion, residual temporal variation after global motion compensation, and weak motion.
TSC-Net produces 0, 1, 5, 1, 0, and 3 false positives, respectively (FPR $\leq 0.20\%$).
TSC-Net yields few false positives even with rich high-frequency content or strong motion.
Subset definitions, per-method TN/FP counts, and weak-motion results are given in Sec.~\ref{app:fpr}.
\end{revblock}

\subsection{Ablation Study}
\begin{table}[!t]
	\revon
	\centering
	\footnotesize
	\setlength{\tabcolsep}{3.2pt}
	\caption{\rev{Compact ablation on slice construction and temporal coverage (mAP / mAUROC). Ours uses one temporal window. ``1-Seg.\ MP'' selects multiple HF/LF patches in that window; ``4-Seg.'' selects one HF/LF patch pair per window across four windows. Full results and runtimes are in Tables~\ref{table:ablation_diff} and~\ref{table:ablation_seg}.}}
	\label{table:ablation_main}
	\begin{tabular}{l cc cc}
		\toprule
		\multirow{2}{*}{\textbf{Method}} & \multicolumn{2}{c}{\textbf{GenVideo}} & \multicolumn{2}{c}{\textbf{AI-Artist}} \\
		\cmidrule(lr){2-3}\cmidrule(lr){4-5}
		& \textbf{mAP} & \textbf{mAUROC} & \textbf{mAP} & \textbf{mAUROC} \\
		\midrule
		Select Patch & 0.9871 & 0.9846 & 0.9644 & 0.9684 \\
		Resize       & 0.9919 & 0.9893 & 0.9541 & 0.9365 \\
		Only Row     & 0.9965 & 0.9958 & 0.9735 & 0.9762 \\
		Only Column  & 0.9969 & 0.9960 & 0.9815 & 0.9784 \\
		\midrule
		1-Seg.\ MP   & 0.9904 & 0.9879 & 0.9808 & 0.9807 \\
		4-Seg.       & \textbf{0.9982} & \textbf{0.9977} & \textbf{0.9849} & \textbf{0.9836} \\
		\midrule
		Ours (1-Seg.) & 0.9973 & 0.9967 & 0.9805 & 0.9773 \\
		\bottomrule
	\end{tabular}
	\vspace{-\intextsep}
\end{table}
\rev{Table~\ref{table:ablation_main} shows that Select Patch (no temporal slices) and Resize (resizing instead of frequency-aware patch localization) both underperform the full model on both benchmarks. Column-only slices perform slightly better on AI-Artist; combining directions performs better on GenVideo. With the same eight-slice budget per directional branch, 4-Seg.\ improves all metrics but increases preprocessing time from 3.198 to 8.908\,ms, while model time stays near 1.5\,ms. We therefore retain the single-window default.} Sec.~\ref{app:ablation} \rev{provides the full} ablations, including \rev{temporal orders, HF/LF groups,} positional encoding, \rev{and hyperparameters. We use} $L=3$, 4 heads, $n_{\text{pick}}=8$, and $T_{\max}=8$ throughout.

\section{Conclusion}

In this paper, we presented a lightweight framework for AI-generated video detection based on frequency-aware temporal slice consistency learning. 
Rather than relying on heavy video backbones or pretrained foundation models, our method reparameterizes videos into compact slice-domain representations that explicitly preserve local temporal evolution while suppressing redundant appearance content. 
We further proposed the Temporal Slice Consistency Network, which integrates task-driven slice localization and assembly, hierarchical positional encoding, and lightweight Transformer-based modeling to capture cross-slice and cross-frequency dependencies. 
Extensive experiments on GenVideo and our newly constructed AI-Artist dataset demonstrate that our method achieves strong overall detection performance while requiring orders-of-magnitude fewer FLOPs than existing video-based detectors.

\clearpage
\begin{revblock}
\begin{ack}
This research is supported by the National Research Foundation Singapore under the AI Singapore Programme (AISG Award No: AISG4-TC-2025-018-SGKR), and the Lee Kong Chian Fellowships. The authors declare no competing interests.
\end{ack}
\end{revblock}

\bibliographystyle{plain_revision}
\small
\bibliography{bibliography_camera_ready,bibliography_camera_ready_extra}

\newpage 
\appendix
\section*{Appendix Contents}
\vspace{0.5em}
\noindent\rule{\linewidth}{0.8pt} %
\vspace{1.2em}

\newcommand{\tocsection}[2]{%
	\noindent\hyperref[#1]{\textbf{\ref*{#1}}\quad#2}\hfill\pageref{#1}\par\vspace{3pt}}
\newcommand{\tocsubsection}[2]{%
	\noindent\hspace{2em}\hyperref[#1]{\ref*{#1}\quad#2}\ \textcolor{black!50}{\dotfill}\ \pageref{#1}\par}

\tocsection{app:related_work}{Related Work}
\tocsubsection{app:image_detection}{AI-Generated Image Detection}
\tocsubsection{app:video_detection}{AI-Generated Video Detection}
\tocsection{app:motivation}{Motivation of Temporal Slice Representation}
\tocsection{app:dataset}{AI-Artist Dataset}
\tocsubsection{app:collection_data}{Collection details of AI-Artist}
\tocsubsection{app:challenge_data}{Diversity and Realism of AI-Artist}
\tocsection{app:ablation}{Ablation Study}
\tocsubsection{app:ablation_hyper}{Ablation on Key Hyperparameters}
\tocsubsection{app:ablation_TS}{Ablation on Temporal Slice Representation}
\tocsubsection{app:ablation_comp}{Ablation on Temporal Slice Construction and Positional Encoding}
\rev{\tocsubsection{app:ablation_dir_seg}{Ablation on Slice Direction and Temporal Coverage}}
\rev{\tocsubsection{app:ablation_loc}{Alternative Slice Localization}}
\rev{\tocsection{app:pipeline_cost}{Full-Pipeline Computational Cost}}
\rev{\tocsection{app:fpr}{False Positives on Real Videos and Weak-Motion Analysis}}
\tocsection{app:feature_visualization}{Feature Space Visualization}
\tocsection{app:limitations}{Limitations and Future Work}
\tocsection{app:impacts}{Potential Societal Impacts}
\tocsection{app:visualization_data}{Visualization of AI-Artist}

\vspace{1em}
\noindent\rule{\linewidth}{0.4pt} %
\vspace{2em}

\section{Related Work}
\label{app:related_work}
\subsection{AI-Generated Image Detection}
\label{app:image_detection}
Existing methods can be broadly categorized into three lines. 
The first line exploits frequency-domain or local low-level forensic artifacts left by generative models. Early works studied spatial-domain inconsistencies such as co-occurrence patterns~\cite{nataraj2019detecting}, and abnormal reflections~\cite{o2012exposing}, while later methods revealed that synthetic images also exhibit characteristic artifacts in frequency spectra and local textures~\cite{wang2020cnn, frank2020leveraging}. Recently, some methods leveraged content-agnostic low-level statistical cues induced by generative models for more generalizable AI-generated image detection \cite{cozzolino2024zero,yuan2025mlep,tan2024rethinking}.
The second line leverages large pretrained models to obtain more transferable representations. UnivFD~\cite{ojha2023towards} builds a universal classifier on pretrained CLIP features, while DIRE~\cite{wang2023dire} exploits reconstruction discrepancies derived from a pretrained diffusion model. Along this direction, Zhang~\etal~\cite{zhang2025towards} argue that pretrained representations often contain substantial task-irrelevant information, and that suppressing such irrelevant components can improve generalization.
The third line combines low-level forensic artifacts with high-level semantic representations. FatFormer~\cite{liu2024forgery} integrates local forgery traces with CLIP-based features, while AIDE~\cite{yan2024sanity} follows a similar motivation by combining semantic information with patch-level frequency cues. 
Overall, image-based detectors have demonstrated promising generalization ability, yet they mainly operate on individual frames and therefore do not explicitly model temporal evolution, which is critical for AI-generated video detection. 
Moreover, many recent methods rely on large pretrained semantic models, resulting in non-trivial computational overhead. 
In contrast, our method directly encodes temporal cues into compact frequency-aware temporal slices, enabling efficient video detection without relying on large pretrained backbones.

\subsection{AI-Generated Video Detection}
\label{app:video_detection}
Existing methods for AI-generated video detection can be broadly categorized into two lines.
The first line learns spatio-temporal forensic inconsistencies with dedicated video detection models, such as TPN~\cite{yang2020temporal} and DeMamba~\cite{chen2024demamba}. 
These methods typically process resized frame sequences using video backbones or temporal modules, and thus inherit the computational cost and redundancy of full-frame modeling.
The second line exploits more transferable temporal cues in pretrained representation space. 
Internò~\etal~\cite{interno2025aigeneratedvideodetectionperceptual} \rev{show} that the geometry of temporal trajectories in pretrained feature space provides a discriminative signal, while D3~\cite{zheng2025d3} demonstrates that training-free second-order temporal statistics can also serve as effective generalized cues. 
Although effective, these methods still rely on frame-level pretrained representations and perform temporal reasoning after feature extraction.
Related to our motivation, temporal inconsistency has also been studied in face forgery detection. 
STIL~\cite{gu2021spatiotemporal} uses slice-like observations to illustrate temporal inconsistency and introduces feature-level temporal difference operations along horizontal and vertical directions within a CNN block. 
Different from such feature-level temporal-difference modeling, our method constructs temporal slices before feature extraction and treats them as the primary input representation for AI-generated video detection. 
Moreover, our slice construction is frequency-aware, explicitly selecting informative high- and low-frequency regions and modeling cross-slice and cross-frequency dependencies with a lightweight Transformer.
In addition to detector design, recent work also emphasizes more realistic and challenging benchmarks. 
GenVideo~\cite{chen2024demamba} and GenVidBench~\cite{ni2026genvidbench} highlight the importance of cross-source and cross-generator evaluation protocols for AI-generated video detection. 
Building on this direction, we further construct AI-Artist to evaluate detector generalization under recent high-fidelity video generators.
Overall, existing video detectors mainly follow the full-frame resize paradigm or feature-space temporal reasoning paradigm. 
While effective in certain settings, such designs often introduce substantial computational overhead and redundant appearance content. 
In contrast, our method reformulates videos into compact frequency-aware temporal slices, where local temporal evolution is explicitly encoded in the input representation for lightweight and effective detection.

\section{Motivation of Temporal Slice Representation}
\label{app:motivation}
\begin{figure}[!hbt]\footnotesize
	\centering
	\includegraphics[width=\linewidth]{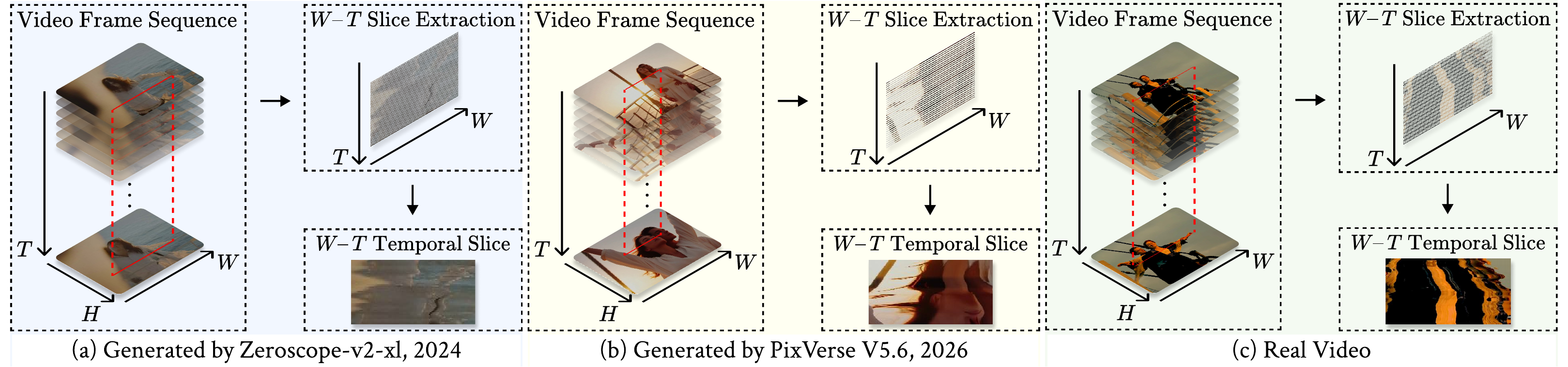} 
	\caption{Visual comparison of temporal slice representations for videos generated by an earlier model, a more recent model, and a real video. (a) A sample generated by Zeroscope-v2-xl~\cite{zeroscope2024}, an earlier text-to-video model included in GenVideo. (b) A sample generated by PixVerse V5.6~\cite{PixVerse2026}, a recent commercial model from our newly collected AI-Artist benchmark. (c) A real video. Following the red dashed line, we unfold one row of pixels along the temporal dimension to obtain the temporal slice. }
	\label{fig:distribution_patch}
\end{figure}

In the main manuscript, Fig.~\ref{fig:motivation} conceptually motivates why temporal slices provide a more task-aligned representation than conventional full-frame or patch-based processing, while Fig.~\ref{fig:intro_abc}(b) further shows that slice-domain statistics yield substantially clearer class separability than patch-domain representations. Here, we complement these analyses with direct visual evidence.

Fig.~\ref{fig:distribution_patch} compares temporal slices extracted from three representative cases: an earlier generator, a recent high-fidelity generator, and a real video. For fair comparison, the generated examples are conditioned on text prompts automatically extracted from the corresponding real video using ChatGPT5.4~\cite{GPT2026}, and then synthesized by the respective video generation models. The sample from the earlier generator exhibits obvious blurring, jagged transitions, and unstable local structures, indicating weak temporal consistency across frames. In contrast, the recent generator produces noticeably cleaner slices with sharper contours and smoother transitions, reflecting the substantial progress of modern video generation models and the increasing difficulty of detection.

Nevertheless, despite these improvements, the generated sample still differs from the real video. The real slice presents cleaner trajectories, more stable structures, and more coherent local variations over time, whereas the sample generated by the recent commercial model still contains abrupt transitions, distorted contours, or irregular fluctuations in certain regions. These residual discrepancies are often subtle in individual frames, but become much more explicit after reorganizing the video into temporal slices.
This comparison further supports the core motivation of our method: Temporal slices explicitly preserve structured temporal evolution while suppressing redundant appearance content, thereby exposing temporal flickering, motion inconsistency, and structural instability in a compact representation. Importantly, such discriminative cues remain observable not only for earlier generators, but also for recent frontier models.

Overall, Fig.~\ref{fig:distribution_patch} provides intuitive visual evidence that is highly consistent with the quantitative findings in Fig.~\ref{fig:intro_abc}(b), and further justifies our design choice of frequency-aware temporal slice modeling.

\section{AI-Artist Dataset}
\label{app:dataset}
\subsection{Collection details of AI-Artist}
\label{app:collection_data}

To construct a challenging benchmark for AI-generated video detection, we collected videos generated by recent high-performing commercial video generation models, together with highly realistic AI-generated videos shared by professional creators on public online platforms. The overall pipeline contains two complementary branches: \textit{web-collection} and \textit{self-generation}. The resulting dataset contains 380 videos in total, including 221 web-collected videos and 159 self-generated videos. The web-collected branch aims to capture realistic AI-generated videos that are naturally disseminated in online communities, while the self-generated branch provides controllable samples with diverse subjects, actions, scenes, and camera motions. Statistics of all collected subsets, including duration, resolution, and frame rate, are summarized in Table~\ref{tab:video_subset_statistics}.

\begin{table}[!t]
	\centering
	\footnotesize
	\caption{Statistics of the AI-generated video subsets in AI-Artist. For binary evaluation, these generated videos are paired with an equal number of real videos sampled from the GenVideo OOD real-video split.}
	\label{tab:video_subset_statistics}
	\resizebox{\linewidth}{!}{
		\begin{tabular}{llcccc}
			\toprule
			\textbf{Collection Protocol} & \textbf{Sub-dataset} & \textbf{\#Videos} & \textbf{Duration} & \textbf{Resolution} & \textbf{Frame Rate} \\
			\midrule
			\multirow{3}{*}{Web-Collection} 
			& Seedance (Film) & 55  & 0.5--15s & $496 \times 864$--$2940 \times 1260$ & 24--60 fps \\
			& Kling 3.0       & 46  & 1--15s   & $480 \times 816$--$1176 \times 1764$   & 24--30 fps \\
			& AI\_artists     & 120 & 1--8s    & $960 \times 544$--$2538 \times 1080$   & 24--30 fps \\
			\midrule
			\multirow{5}{*}{Self-Generation} 
			& Runway Gen-4.5  & 40  & 5s       & 720p                         & 24 fps \\
			& PixVerse V5.6   & 40  & 3s       & 720p                         & 24 fps \\
			& Hailuo 2.3      & 35  & 5s       & $1366 \times 768$            & 24 fps \\
			& Seedance 2.0    & 34  & 1--3s    & $1118 \times 720$            & 24 fps \\
			& Luma Ray3.14    & 10  & 5s       & 720p                         & 24 fps \\
			\bottomrule
		\end{tabular}
	}
\end{table}
\paragraph{Web-collected videos.}

For the web-collected branch, we gathered AI-generated videos from public creator communities and social platforms. Specifically, Seedance (Film) videos contain 55 videos collected from X/Twitter, Discord, Reddit, and related public platforms by searching model-specific keywords and examining the accompanying descriptions, hashtags, or community context. Kling 3.0 videos contain 46 videos collected from its official showcase pages and Discord community, where users publicly share generated results. AI\_artists videos contain 120 videos collected from Douyin using relevant AI-creation keywords.
Since online videos may include irrelevant content, advertisements, tutorials, reposted materials, or samples with ambiguous provenance, we manually reviewed all candidates and retained only videos that were visually valid and could be reasonably associated with the claimed generation source. This branch therefore reflects realistic in-the-wild distributions of AI-generated videos shared by real users and content creators.

\paragraph{Self-generated videos.}
For the self-generated branch, we generated videos using five commercial models: Runway Gen-4.5~\cite{Runway2025}, PixVerse V5.6~\cite{PixVerse2026}, Hailuo 2.3~\cite{Hailuo2025}, Seedance 2.0~\cite{seedance2026seedance}, and Luma Ray3.14~\cite{Luma2026}, resulting in 40, 40, 35, 34, and 10 videos, respectively. All prompts were constructed with the assistance of a large language model. 
For Seedance 2.0, we focused on human-centric videos with diverse identities, actions, and daily-life scenarios. The subject pool includes people of different ages, genders, and appearances, such as \textit{a little girl}, \textit{an old couple}, \textit{a middle-aged man}, and \textit{a young chef}. These subjects were paired with common human activities, including \textit{watching TV}, \textit{doing yoga}, \textit{walking in the street}, \textit{cooking}, and \textit{talking with friends}. We further diversified the environments with both indoor and outdoor scenes, such as \textit{at home}, \textit{in the kitchen}, \textit{on the playground}, \textit{in a park}, \textit{in a classroom}, and \textit{on a city street}. This design encourages rich human motion patterns, diverse backgrounds, and realistic temporal variations.
For the remaining four commercial models, including Runway Gen-4.5, PixVerse V5.6, Hailuo 2.3, and Luma Ray3.14, we adopted a more systematic prompt construction strategy inspired by GenVideo~\cite{chen2024demamba}. 
We first built a subject dictionary containing 50 diverse subjects spanning humans, animals, vehicles, objects, and natural scenes. Representative examples include \textit{a yellow school bus}, \textit{a flowing waterfall}, \textit{a striped orange tiger}, \textit{a young dancer in a red skirt}, \textit{a white sports car}, and \textit{a small rabbit}. For each subject, we designed multiple plausible actions and scene contexts to ensure semantic diversity. In addition, we introduced 20 camera motion patterns, such as \textit{fast tracking shot}, \textit{slow push-in shot}, \textit{low-angle rising shot}, \textit{drone fly-through shot}, and \textit{handheld follow shot}. These elements were then combined and expanded by a large language model into 300 concise and visually clear prompts, which were submitted to the official websites of the corresponding models under default generation settings.

\paragraph{Real-video reference set for evaluation.}
AI-Artist is designed to focus on recent AI-generated videos and therefore contains only generated samples. To compute binary detection metrics, we pair AI-Artist with real videos sampled from the real-video split of the GenVideo OOD test set. Specifically, we sample 380 real videos to match the number of AI-Artist generated videos, resulting in a balanced evaluation set. The sampled real subset is fixed and shared across all compared detectors. For per-generator evaluation, the same protocol is applied by matching the number of real videos to the corresponding generated subset. This avoids changes in class prior and ensures that AP, AUROC, and ACC are computed under a consistent evaluation setting.

\paragraph{Summary.}
Overall, our collection strategy combines in-the-wild online videos with controllable model-generated videos. The web-collected subsets provide realistic distribution diversity by reflecting AI-generated videos shared across public platforms, while the self-generated subsets ensure broad semantic coverage over subjects, actions, scenes, and camera motions. As a result, AI-Artist covers both realistic dissemination scenarios and systematically generated samples from recent frontier commercial models.

\subsection{Diversity and Realism of AI-Artist}
\label{app:challenge_data}
\begin{wrapfigure}{R}{0.67\textwidth} 
	\footnotesize
	\centering
	\includegraphics[width=\linewidth]{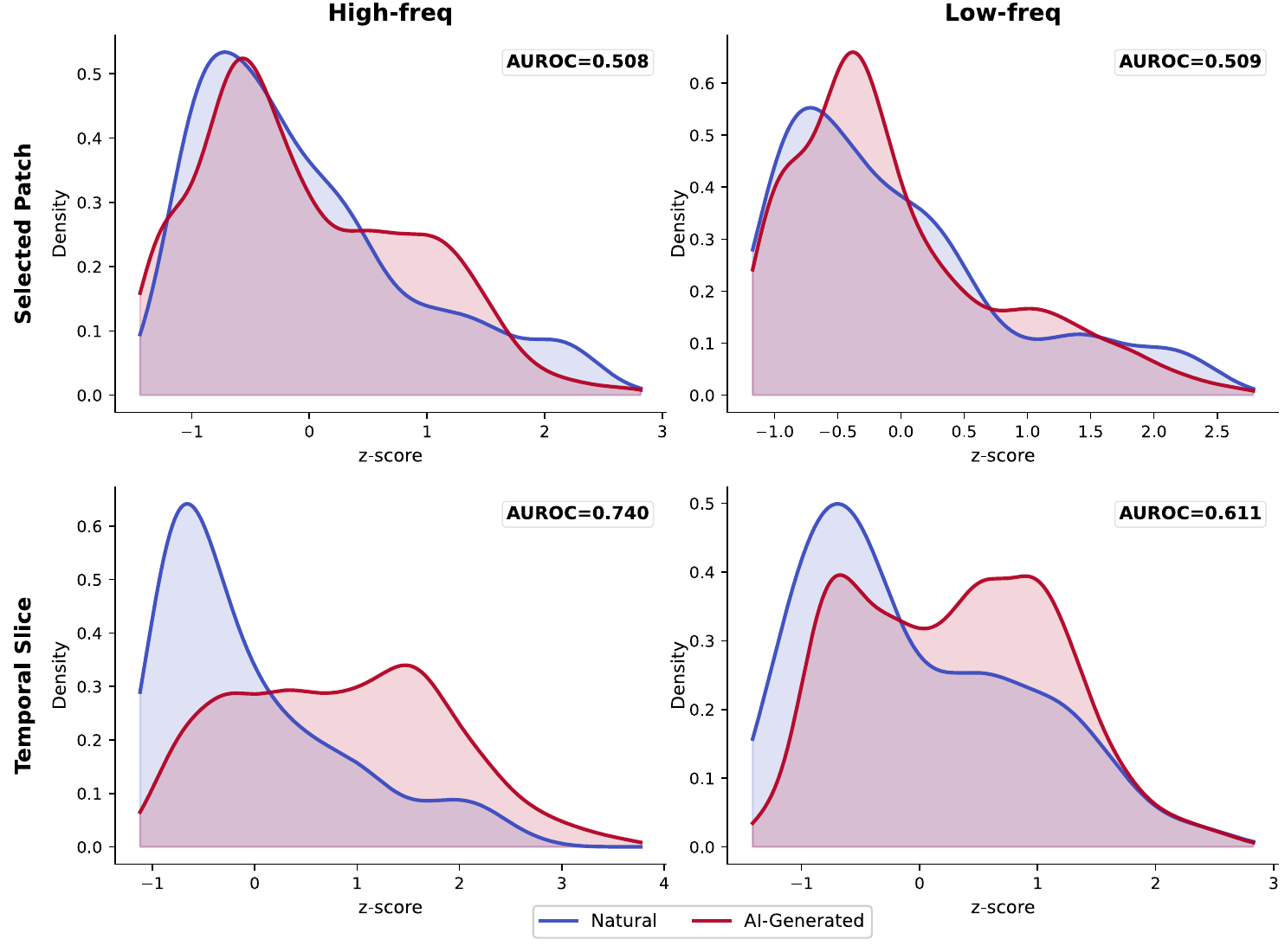} 
	\caption{Distributional analysis of natural (\textcolor{blue}{blue}) videos in GenVideo and AI-generated (\textcolor{red}{red}) videos in AI-Artist under patch-domain and slice-domain representations. The top row shows selected spatial patches, while the bottom row shows the corresponding temporal slices. Columns correspond to high-frequency and low-frequency spatial regions. Compared with Fig.~\ref{fig:intro_abc}(b), AI-Artist exhibits stronger overlap between natural and AI-generated distributions, indicating that recent high-fidelity generators substantially reduce simple distributional discrepancies.}
	\label{fig:distribution_AIArt}
	\vspace{-\intextsep} %
\end{wrapfigure}
Modern video generation models have rapidly advanced in both visual fidelity and temporal coherence, raising a critical question for detection: \emph{what makes AI-generated videos still distinguishable when their observable statistics become increasingly similar to real data?}

To investigate this, we extend the distributional analysis in Fig.~\ref{fig:intro_abc}(b) to the AI-Artist evaluation setting, where AI-generated videos come from AI-Artist and natural videos are sampled from the GenVideo OOD real-video split following the balanced protocol described in Sec.~\ref{app:collection_data}. As shown in Fig.~\ref{fig:distribution_AIArt}, the selected patch distributions of natural and AI-generated videos are almost indistinguishable in both high-frequency and low-frequency regions, with AUROC values close to random guessing. This indicates that local patch-level statistics provide limited discriminative evidence on AI-Artist.
After reorganizing the same content into temporal slices, the separability between natural and AI-generated videos becomes more evident. In particular, high-frequency temporal slices show a clearer distributional gap, suggesting that temporal slice representations can still expose structured temporal differences that are largely hidden in isolated spatial patches. Nevertheless, compared with Fig.~\ref{fig:intro_abc}(b), the slice-domain separability on AI-Artist is reduced, especially in low-frequency regions. This reflects the increased realism and temporal coherence of recent generation models, which makes AI-Artist more challenging than earlier benchmarks.
\rev{This observation suggests that detection becomes harder as simple statistical differences diminish.} Earlier generated videos often contain more obvious low-level artifacts or unstable temporal patterns, making them easier to distinguish using simple statistical cues. In contrast, recent generators substantially narrow the distributional gap to real videos, particularly at the patch level. Therefore, \rev{these results motivate using structured temporal cues alongside} local statistics.
Motivated by this insight, we construct the AI-Artist dataset with two key objectives. First, it emphasizes \emph{diversity}, incorporating outputs from multiple state-of-the-art generation systems with different architectures and generation behaviors. This ensures that detectors are evaluated under a broad range of unseen generation patterns. Second, it emphasizes \emph{realism}, focusing on recent high-quality videos where conventional artifacts are substantially reduced. Together, these factors make AI-Artist a more challenging and realistic benchmark for evaluating generalization.

The impact of this increased difficulty is reflected in Table~\ref{table:AI_Artist}. Many existing methods experience noticeable performance degradation, especially those relying heavily on local or low-level statistical cues. Even strong baselines exhibit instability across generators, indicating limited robustness when the distributional gap between real and generated videos becomes smaller.
In contrast, our method achieves the best average AP and AUROC and remains among the top-performing methods on most generator subsets. These results suggest that our approach remains effective under reduced patch-level separability. By modeling cross-slice temporal dynamics and cross-frequency interactions, our method focuses on structured temporal patterns that remain informative even for high-fidelity generated videos.

Overall, AI-Artist not only provides a more stringent evaluation benchmark, but also highlights a key trend in AI-generated video detection: as patch-level distributional differences diminish, effective detection increasingly depends on modeling structured temporal slice representations rather than relying on isolated spatial statistics.

\begin{table}[!t]
	\caption{\rev{Comparison with state-of-the-art methods in terms of} \rev{Accuracy} (ACC), \rev{Average Precision} (AP) and Area Under the Receiver Operating Characteristic Curve (AUROC) on the AI-Artist dataset. }
	\label{table:AI_Artist}
	\centering
	\footnotesize
	\setlength{\tabcolsep}{3.4pt}
	\begin{tabular}{l l *{8}{c} c}
		\toprule
		\multirow{4}{*}{\textbf{Model}} &
		\multirow{4}{*}{\textbf{Metric}} &
		\multicolumn{8}{c}{\textbf{AI-Artist}} &
		\multirow{4}{*}{\textbf{Mean}} \\
		\cmidrule(lr){3-10}
		& & \makecell{\textbf{Seedance}\\\textbf{(Film)}}  & \textbf{Seedance} & \textbf{Kling} & \textbf{Runway} & \textbf{PixVerse} & \textbf{Hailuo} & \textbf{Luma} & \textbf{AI\_artists} & \\
		\midrule
		\multirow{3}{*}{MLEP (I)}
		& ACC   & 0.0926 & 0.0000 & 0.1522 & 0.1000 & 0.1795 & 0.0000 & 0.0000 & 0.0083 & 0.0666 \\
		& AP    & 0.4942 & 0.5377 & 0.5361 & 0.5544 & 0.6382 & 0.5345 & 0.6031 & 0.3387 & 0.5296  \\
		& AUROC & 0.4781 & 0.6168 & 0.4920 & 0.5262 & 0.6220 & 0.4971 & 0.5900 & 0.1622 & 0.4980  \\
		\midrule
		\multirow{3}{*}{NPR (I)}
		& ACC   & 0.4444 & 0.1471 & 0.8696 & \textbf{1.0000} & \textbf{1.0000} & \textbf{1.0000} & \textbf{1.0000} & 0.3750 & 0.7295 \\
		& AP    & 0.7717 & 0.6426 & 0.9342 & 0.9882 & 0.9929 & 0.9949 & \textbf{1.0000} & 0.6768 & 0.8752 \\
		& AUROC & 0.7383 & 0.7232 & 0.9258 & 0.9888 & 0.9934 & 0.9951 & \textbf{1.0000} & 0.6296 & 0.8743 \\
		\midrule
		\multirow{3}{*}{VIBAIGCD (I)}
		& ACC   &  0.6296 & 0.5588 & 0.6087 & 0.9000 & 0.7692 & 0.7143 & 0.7000 & 0.6417 & 0.6903 \\
		& AP    & 0.6522 & 0.6406 & 0.6157 & 0.7902 & 0.7517 & 0.7360 & 0.6782 & 0.5964 & 0.6826 \\
		& AUROC & 0.6667 & 0.6384 & 0.6181 & 0.8575 & 0.7495 & 0.7208 & 0.7400 & 0.6285 & 0.7024 \\
		\midrule
		\multirow{3}{*}{AIDE (I)}
		& ACC   & 0.5000 & 0.1765 & 0.8478 & 0.9000 & 0.7692 & \textbf{1.0000} & \textbf{1.0000} & 0.3667 & 0.6950 \\
		& AP    & 0.7880 & 0.5960 & 0.9537 & 0.9503 & 0.8885 & \textbf{0.9985} & 0.9063 & 0.7298 & 0.8514 \\
		& AUROC & 0.7716 & 0.5856 & 0.9504 & 0.9531 & 0.8915 & \textbf{0.9984} & 0.9300 & 0.7189 & 0.8499 \\
		
		\midrule
		\multirow{3}{*}{TALL (V)}
		& ACC   & \textbf{0.6852} & 0.8529 & \textbf{0.9130} & \textbf{1.0000} & 0.9231 & 0.9429 & \textbf{1.0000} & \textbf{0.7167} & \textbf{0.8792} \\
		& AP    & 0.6119 & 0.6071 & 0.8252 & 0.9414 & 0.8481 & 0.9367 & 0.7896 & 0.5869 & 0.7684  \\
		& AUROC &  0.5724 & 0.5484 & 0.7765 & 0.9394 & 0.8087 & 0.9086 & 0.7700 & 0.5301 & 0.7318   \\
		\midrule
		
		\multirow{3}{*}{XCLIP (V)}
		& ACC   & 0.3333 & 0.2059 & 0.6304 & 0.9750 & 0.9487 & \textbf{1.0000} & \textbf{1.0000} & 0.2250 & 0.6648  \\
		& AP    & 0.3742 & 0.3478 & 0.5109 & 0.6927 & 0.7013 & {0.7965} & 0.4733 & 0.3489 & 0.5307   \\
		& AUROC & 0.2853 & 0.1799 & 0.5118 & 0.7313 & 0.7094 & {0.8122} & 0.4600 & 0.2137 & 0.4880   \\
		\midrule
		\multirow{3}{*}{DeMamba (V)}
		& ACC   & {0.6667}  & 0.4118  & 0.8913  & 0.9250 & {0.9487} & \textbf{1.0000} & \textbf{1.0000} & 0.2833 & {0.7658}  \\
		& AP    & 0.4276 & 0.3728 & 0.6274  & 0.6334  & 0.6758  & 0.9747 & 0.5026 & 0.3371 &   0.5689 \\
		& AUROC & 0.3755 & 0.2673 & 0.6129 & 0.6425 & 0.6785 & 0.9796 & 0.5300 & 0.1689 & 0.5319 \\
		\midrule
		\multirow{3}{*}{D3 (V)}
		& ACC   & --- & --- & --- & --- & --- & --- & --- & --- &  --- \\
		& AP    & \textbf{0.9292} & 0.9874 &  0.9728  & 0.9922 & 0.9927 & 0.9929  & \textbf{1.0000} & 0.9317   &  0.9749  \\
		& AUROC & 0.9150 & 0.9827 & 0.9683 & 0.9906 & 0.9915 &  0.9918  &  \textbf{1.0000}  & 0.9151   &   0.9694  \\
		\midrule
		\multirow{3}{*}{ReStraV (V)}
		& ACC   & 0.1296  & \textbf{0.9706} & 0.7826 & 0.9500  & \textbf{1.0000} & 0.9143 & \textbf{1.0000} & 0.5667 &  0.7892 \\
		& AP    & {0.9261} & 0.9820 & 0.9780 & 0.9809 & 0.9782 & 0.9777  & 0.9782 & 0.9656 &   0.9712 \\
		& AUROC & \textbf{0.9207}  & 0.9811 & 0.9787 & 0.9822 & 0.9790 & 0.9776  & 0.9888 & 0.9549 &   0.9704 \\
		\midrule
		
		\multirow{3}{*}{Ours (V)}
		& ACC   & {0.3333} & 0.9118 & \textbf{0.9130} & \textbf{1.0000} & \textbf{1.0000} & 0.9714 & \textbf{1.0000} & \textbf{0.7167} &   {0.8558} \\
		& AP    & {0.8883} & \textbf{0.9933} & \textbf{0.9958} & \textbf{1.0000} & \textbf{1.0000} & {0.9976} & \textbf{1.0000} & \textbf{0.9692} & \textbf{0.9805} \\
		& AUROC & {0.8735} & \textbf{0.9931} & \textbf{0.9957} & \textbf{1.0000} & \textbf{1.0000} & {0.9976} & \textbf{1.0000} & \textbf{0.9587} &   \textbf{0.9773} \\
		\bottomrule
	\end{tabular}
\end{table}

\section{Ablation Study}
\label{app:ablation}
\begin{figure}[!hbt]\footnotesize
	\centering
\includegraphics[width=\linewidth]{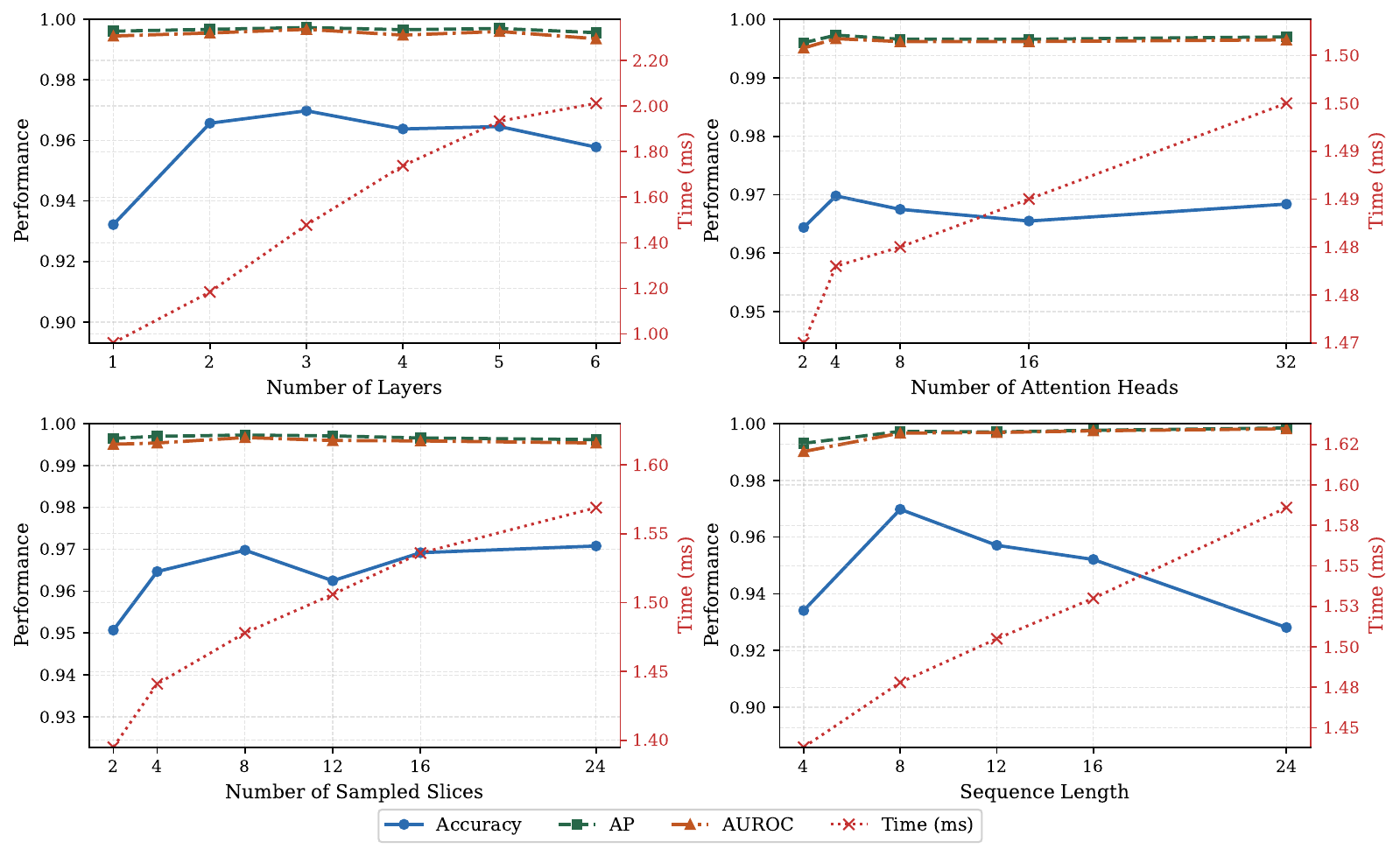}
\vspace{-1.5em}
\captionof{figure}{%
	Ablation on key hyperparameters.
}
\label{fig:ablation}
\vspace{-\intextsep} %
\end{figure}
\subsection{Ablation on Key Hyperparameters}
\label{app:ablation_hyper}
We first analyze the sensitivity of key hyperparameters in TSC-Net. 
As shown in Fig.~\ref{fig:ablation}, we vary four important factors, including the number of Transformer layers, the number of attention heads, the number of sampled slices, and the temporal window length.
Overall, the performance remains stable across a broad range of settings, especially in terms of mAP and mAUROC, indicating that our framework is not overly sensitive to hyperparameter choices. 
For the number of Transformer layers, performance improves as the model becomes deeper and reaches a favorable trade-off at 3 layers, while further increasing the depth brings no clear additional gain. 
A similar trend is observed for the number of attention heads, where 4 heads are sufficient to achieve strong performance. 
For the number of sampled slices, the best overall results are obtained with $n_{\text{pick}}=8$, which we therefore adopt as the default setting. 
For the temporal window length, mAP and mAUROC slightly improve as the sequence becomes longer, suggesting that longer temporal context can provide useful information. 
However, ACC decreases when the window length continues to increase, possibly because longer windows introduce more padding for short videos, which can affect threshold-dependent accuracy. 
Considering both overall performance and fair comparison with prior video-based baselines, we use $T_{\max}=8$ in all experiments.

\subsection{Ablation on Temporal Slice Representation}
\label{app:ablation_TS}
\begin{table}[!t]
	\caption{Ablation study on different temporal slice representations, including temporal-order slice representations, slice construction strategies, \rev{slice directions,} high- and low-frequency components, and positional encoding variants.  Results are reported on GenVideo and AI-Artist in terms of ACC, mAP, and mAUROC. Inference time (ms) is also reported.}
	\label{table:ablation_diff}
	\centering
	\small
\begin{tabular}{l c c c c c c c}
	\toprule
	\multirow{2}{*}{\textbf{Method}} & \multicolumn{3}{c}{\textbf{GenVideo}} & \multicolumn{3}{c}{\textbf{AI-Artist}}  & \textbf{Time} \\
	\cmidrule(lr){2-7}
	&  {\textbf{ACC}}  &  {\textbf{mAP}} &  {\textbf{mAUROC}} &  {\textbf{ACC}}  &  {\textbf{mAP}} &  {\textbf{mAUROC}} &  \textbf{(ms)}  \\
	\midrule
	Raw Temporal Slice    & 0.9651  & 0.9970  & 0.9962  & 0.8618  &  0.9775 &  0.9763  & {1.426} \\
	1st-order Diff Slice       & 0.9658    & 0.9969  & 0.9955  & \textbf{0.8638} & 0.9745  &  0.9757  & 1.460 \\
	2nd-order Diff Slice   & 0.9635    & 0.9939  & 0.9906  & 0.8520  &  0.9691 & 0.9562  & 1.465 \\
	\midrule
	Select Patch    & 0.9164 & 0.9871  & 0.9846   & 0.8536 & 0.9644 &  0.9684& 1.414 \\
	Resize       & 0.9323    & 0.9919   & 0.9893   &  0.8297 & 0.9541 &   0.9365 & 1.463 \\
	Only HF   & 0.9503   & 0.9947 & 0.9912  & 0.8522 & 0.9723 &  0.9572  & 1.436 \\
	Only LF       & 0.9336 & 0.9943  & 0.9909   & {0.8615} &  \textbf{0.9816} & \textbf{0.9795} & 1.437 \\
	Fixed PE   & 0.9522  & 0.9961 & 0.9940   &  0.8602  &  0.9796 &  0.9752  & \textbf{1.355} \\
	Learnable PE   & 0.9494  & 0.9948  & 0.9910    & 0.8565  & 0.9745 &  0.9652 &  1.445 \\
		\midrule
		\rev{Only Row} & \rev{0.9664} & \rev{0.9965} & \rev{0.9958} & \rev{0.8526} & \rev{0.9735} & \rev{0.9762} & \rev{1.458} \\
		\rev{Only Column} & \rev{0.9668} & \rev{0.9969} & \rev{0.9960} & \rev{0.8592} & \rev{0.9815} & \rev{0.9784} & \rev{1.459} \\
	\midrule
	Ours       & \textbf{0.9698}   & \textbf{0.9973}  & \textbf{0.9967}  & 0.8558 &  {0.9805}  & {0.9773} & {1.478} \\
	\bottomrule
\end{tabular}
\vspace{-\intextsep}
\end{table}
We investigate the impact of different temporal-order representations, including raw temporal slices, first-order differences, and second-order differences. 
The results are reported in the upper block of Table~\ref{table:ablation_diff}. 
Overall, no single temporal order consistently dominates across all datasets and metrics, indicating that they capture different aspects of temporal inconsistency.

Raw temporal slices achieve strong performance on both datasets, reaching 0.9651 ACC on GenVideo and 0.8618 ACC on AI-Artist. 
This shows that direct slice representations already preserve useful spatiotemporal consistency cues. 
First-order difference slices further emphasize temporal variations and achieve the highest ACC on AI-Artist among the temporal-order variants, suggesting their effectiveness in capturing motion-level inconsistencies under more challenging generation conditions. 
Second-order difference slices model higher-order temporal dynamics, but their standalone performance is weaker, especially on AI-Artist. 
This is likely because second-order differences amplify not only temporal discrepancies but also noise and local instability, making the resulting representation less reliable when used alone.

By integrating raw slices, first-order differences, and second-order differences, our full model achieves the best performance on GenVideo and remains highly competitive on AI-Artist. 
In particular, it reaches 0.9698 ACC, 0.9973 mAP, and 0.9967 mAUROC on GenVideo, outperforming all single-order variants. 
These results suggest that different temporal orders provide complementary information, and their joint modeling improves the overall robustness of AI-generated video detection. 
Moreover, the inference time remains comparable across variants, indicating that the gain brought by multi-order modeling does not introduce substantial computational overhead.

\noindent\textbf{Bridging Distributional Evidence and Empirical Performance.}
At first glance, the clear distributional discrepancy observed in temporal-difference domains in Fig.~\ref{fig:intro_abc}(b) and Fig.~\ref{fig:distribution_AIArt} may seem inconsistent with the relatively weaker standalone performance of second-order difference slices in Table~\ref{table:ablation_diff}. 
This is not contradictory. 
The distributional analysis is intended as a diagnostic visualization of marginal slice-response statistics, rather than a direct surrogate for end-to-end detection performance. 
Although second-order differences can amplify real--fake discrepancies, they may also amplify noise and discard stable structural information that is useful for robust classification. 
As a result, second-order slices can provide informative evidence while being less reliable as the only input representation. 
In contrast, raw temporal slices preserve stable structural cues, and first-order differences capture local motion variations. 
This motivates our multi-order design: instead of relying on a single temporal signal, we jointly model raw, first-order, and second-order slice representations to exploit their complementary strengths.

\subsection{Ablation on Temporal Slice Construction and Positional Encoding}
\label{app:ablation_comp}
We further validate the effectiveness of temporal slice construction, frequency-aware component design, and positional encoding. 
The results are reported in the lower block of Table~\ref{table:ablation_diff}.

\noindent\textbf{Temporal Slice Construction.}

``Select Patch'' selects the same number of informative patches but does not construct temporal slices. 
Its clear performance drop on both datasets demonstrates that structured temporal slice construction is crucial for exposing temporal inconsistencies, rather than merely selecting discriminative spatial regions. 
``Resize'' directly resizes the windowed video content instead of performing frequency-aware patch localization. 
Its inferior performance suggests that explicitly focusing on localized frequency-aware regions is more effective than relying on globally resized content, where subtle temporal artifacts may be diluted by redundant appearance information.

\noindent\textbf{High- vs. Low-Frequency Temporal Slices.}

We compare variants using only high-frequency patches (``Only HF'') or only low-frequency patches (``Only LF''). 
The results reveal different behaviors across benchmarks. 
On AI-Artist, ``Only LF'' achieves the highest mAP and mAUROC among the ablation variants, reaching 0.9816 and 0.9795, respectively. 
This suggests that, for recent high-fidelity generators, discriminative cues may be more strongly reflected in coarse structural evolution and low-frequency temporal consistency, while high-frequency texture artifacts alone become less reliable.

However, relying solely on low-frequency slices leads to a substantial drop on GenVideo, where ``Only LF'' obtains 0.9336 ACC compared with 0.9698 of the full model. 
This indicates that high-frequency cues remain important for generators that still exhibit local texture instability, boundary flickering, or fine-detail artifacts. 
Conversely, ``Only HF'' also underperforms the full model, especially on AI-Artist, showing that high-frequency information alone is insufficient for detecting more realistic generated videos. 
Therefore, high- and low-frequency temporal slices are complementary: high-frequency slices capture fine-grained local artifacts, while low-frequency slices preserve coarse structural and motion-related inconsistencies. 
Our full model combines both types of cues, achieving the best overall cross-benchmark trade-off.

\noindent\textbf{Positional Encoding.}
Finally, we compare different positional encoding variants. 
The fixed positional encoding preserves deterministic structural coordinates, while the learnable positional encoding allows more flexible adaptation. 
Both variants achieve reasonable performance, but neither consistently matches the full design. 
Our hierarchical positional encoding, which combines frozen structural encoding with a learnable frequency embedding, achieves the best results on GenVideo and strong threshold-free performance on AI-Artist. 
This supports the importance of disentangling structural slice coordinates from frequency-group identity when modeling cross-slice and cross-frequency dependencies.

Overall, these ablation results consistently support the main design choices of TSC-Net: Temporal slices are more effective than patch-only or resizing-based alternatives, multi-order temporal representations provide complementary cues, high- and low-frequency components improve cross-generator robustness, and hierarchical positional encoding benefits structured slice-domain modeling.

\begin{revblock}
\subsection{Ablation on Slice Direction and Temporal Coverage}
\label{app:ablation_dir_seg}
\noindent\textbf{Row-wise vs.\ column-wise slices.}
We evaluate variants that keep only the row-wise branch (``Only Row'') or only the column-wise branch (``Only Column''), with results listed in Table~\ref{table:ablation_diff}.
Both directions are individually effective.
``Only Column'' is slightly stronger on AI-Artist (0.8592/0.9815/0.9784 ACC/mAP/mAUROC versus 0.8558/0.9805/0.9773 for the full model), whereas the joint design is stronger on all GenVideo metrics (0.9698/0.9973/0.9967 versus 0.9664/0.9965/0.9958 for ``Only Row'' and 0.9668/0.9969/0.9960 for ``Only Column'').
This suggests that the preferred slice direction depends on the dataset and that neither direction is consistently better.
Using both directions avoids choosing a direction based on the test distribution and is a more reliable default across benchmarks, with a negligible increase in model time (1.478 versus about 1.46\,ms).

\noindent\textbf{Temporal coverage.}
The default design extracts a single temporal window centered at the key frame.
To study broader temporal coverage, we keep the scoring rules of Eqs.~\eqref{eq:frame_diff} and~\eqref{eq:patch_sel} and change only how the fixed slice budget, four HF and four LF slices in each directional branch, is distributed.
``1-Seg.\ MP'' constructs one window around the key frame $t^\star$, retains the four highest-energy HF patches and the four lowest-energy LF patches according to $S(r,c)$, and assembles their row and column slices.
``4-Seg.'' ranks frame transitions by $d_t$, applies temporal non-maximum suppression to remove adjacent peaks, and selects the top four key frames; a temporal window is constructed around each key frame, and the original localization rule selects one HF/LF patch pair from each window.
All configurations therefore share the same model input shape and differ only in how evidence is sampled.
As shown in Table~\ref{table:ablation_seg}, 4-Seg.\ outperforms both 1-Seg.\ MP and the single-window default on all six metrics, indicating that broader temporal coverage provides complementary information that is not captured by denser spatial sampling within a single window.
The model time remains essentially unchanged, whereas the preprocessing time increases from 3.198 to 8.908\,ms because more frame windows must be accessed and their slices constructed separately.
The single-segment design therefore remains the simpler and more efficient operating point, while 4-Seg.\ is favorable when accuracy is prioritized.

\begin{table}[!t]
	\revon
	\centering
	\small
	\caption{\rev{Ablation on temporal coverage under a matched budget of four HF and four LF slices per directional branch. ``Prep.'' and ``Model'' denote preprocessing and model-forward time per video.}}
	\label{table:ablation_seg}
	\begin{tabular}{l c c c c c c c c}
		\toprule
		\multirow{2}{*}{\textbf{Method}} & \multicolumn{3}{c}{\textbf{GenVideo}} & \multicolumn{3}{c}{\textbf{AI-Artist}} & \textbf{Prep.} & \textbf{Model} \\
		\cmidrule(lr){2-4}\cmidrule(lr){5-7}
		& \textbf{ACC} & \textbf{mAP} & \textbf{mAUROC} & \textbf{ACC} & \textbf{mAP} & \textbf{mAUROC} & \textbf{(ms)} & \textbf{(ms)} \\
		\midrule
		1-Seg.\ MP & 0.9375 & 0.9904 & 0.9879 & 0.8903 & 0.9808 & 0.9807 & 4.668 & \textbf{1.451} \\
		4-Seg. & \textbf{0.9785} & \textbf{0.9982} & \textbf{0.9977} & \textbf{0.9125} & \textbf{0.9849} & \textbf{0.9836} & 8.908 & 1.522 \\
		Ours (1-Seg.) & 0.9698 & 0.9973 & 0.9967 & 0.8558 & 0.9805 & 0.9773 & \textbf{3.198} & 1.478 \\
		\bottomrule
	\end{tabular}
	\vspace{-\intextsep}
\end{table}

\subsection{Alternative Slice Localization}
\label{app:ablation_loc}
Our patch localization relies on a fixed gradient-energy heuristic.
As an alternative, we evaluate a semantically guided variant that localizes regions with Grad-CAM~\cite{selvaraju2017gradcam} from a trained semantic detector before constructing temporal slices.
On GenVideo, this variant obtains 0.9501/0.9961/0.9952 ACC/mAP/mAUROC, compared with 0.9698/0.9973/0.9967 for our gradient-based design, while the parameters, computation, and model latency increase from 1.731\,M/0.109\,GFLOPs/1.478\,ms to 2.810\,M/0.597\,GFLOPs/17.4\,ms.
This indicates that a region important to a semantic classifier is not necessarily where temporal inconsistency is most visible; moreover, Grad-CAM requires an additional model and gradient extraction, which substantially weakens efficiency.
A fully learnable localizer is also not a drop-in replacement in our formulation, because TSC-Net consumes temporal slices rather than the selected patches themselves.
After HF/LF patch localization and discrete cropping, slice construction computes the temporal-difference map and performs a second discrete top-$K$ selection of rows and columns.
Even with a differentiable attention map, the classification loss therefore cannot be back-propagated through these discrete selection steps to the localizer.
End-to-end optimization would require relaxing the discrete crop and row/column selection, or introducing surrogate supervision. These changes could alter the representation and require scoring additional candidates, so the same computational cost cannot be guaranteed.
A lightweight selector designed specifically for slice-level temporal inconsistency is an interesting direction for future work.

\section{Full-Pipeline Computational Cost}
\label{app:pipeline_cost}
\begin{table}[!t]
	\revon
	\centering
	\footnotesize
	\caption{\rev{Stage-wise computational cost of TSC-Net per video on GenVideo.}}
	\label{table:prep_breakdown}
	\begin{tabular}{l c c}
		\toprule
		\textbf{Stage} & \textbf{Time (ms)} & \textbf{GFLOPs} \\
		\midrule
		Temporal-window/key-frame selection and grayscale processing & 0.796 & 0.0415 \\
		Frequency-aware patch selection & 1.652 & 0.0230 \\
		Temporal-slice and difference construction & 0.750 & 0.0136 \\
		\midrule
		Preprocessing subtotal & 3.198 & 0.0781 \\
		Model forward & 1.478 & 0.109 \\
		\midrule
		Full pipeline & 4.676 & 0.187 \\
		\bottomrule
	\end{tabular}
	\vspace{-\intextsep}
\end{table}
Table~\ref{table:complexity} uses the same measurement protocol for all compared methods, separating the model forward pass from method-specific data preprocessing.
For all methods, preprocessing consists only of deterministic, non-learnable operations and introduces no trainable parameters; hence the parameter counts are unaffected.
For TSC-Net, preprocessing comprises temporal-window/key-frame selection with grayscale processing, frequency-aware patch selection, and temporal-slice and difference construction, whose stage-wise costs are listed in Table~\ref{table:prep_breakdown}.

The ``400$\times$ fewer FLOPs'' statement refers to model-forward computation measured under the same protocol for all methods: ReStraV requires 44.20\,GFLOPs, whereas TSC-Net requires 0.109\,GFLOPs, corresponding to 405.5$\times$ fewer model-forward FLOPs; it does not imply a 400$\times$ reduction in end-to-end latency.
FLOPs and wall-clock latency are not expected to scale proportionally, especially in the low-computation regime.
At low computational loads, memory access and kernel launches can account for a larger share of latency, and GPU resources may be underutilized.
Consequently, the 405.5$\times$ reduction in model-forward FLOPs corresponds to a 3.6$\times$ model-forward speedup (1.478 versus 5.307\,ms).
When method-specific preprocessing is included, TSC-Net retains 236$\times$ fewer total GFLOPs (0.187 versus 44.212) and a 2.9$\times$ full-pipeline speedup (4.676 versus 13.444\,ms) over ReStraV, and it has the lowest total GFLOPs and latency among all compared methods.

\section{False Positives on Real Videos and Weak-Motion Analysis}
\label{app:fpr}
\begin{table}[!t]
	\revon
	\centering
	\footnotesize
	\caption{\rev{False-positive analysis on 10,000 real MSR-VTT clips at a threshold of $0.5$, with generated videos as the positive class. Since only real videos are evaluated, only TN and FP are reported. D3 is omitted because it produces an uncalibrated ranking score without a predefined binary threshold.}}
	\label{table:fpr_full}
	\begin{tabular}{l r r c c}
		\toprule
		\textbf{Method} & \textbf{TN} & \textbf{FP} & \textbf{FPR (\%)} & \textbf{Specificity (\%)} \\
		\midrule
		MLEP (I) & 9,578 & 422 & 4.22 & 95.78 \\
		NPR (I) & 8,613 & 1,387 & 13.87 & 86.13 \\
		VIBAIGCD (I) & 5,187 & 4,813 & 48.13 & 51.87 \\
		AIDE (I) & 9,028 & 972 & 9.72 & 90.28 \\
		\midrule
		TALL (V) & 3,051 & 6,949 & 69.49 & 30.51 \\
		XCLIP (V) & 3,494 & 6,506 & 65.06 & 34.94 \\
		DeMamba (V) & 2,609 & 7,391 & 73.91 & 26.09 \\
		ReStraV (V) & 9,961 & 39 & 0.39 & 99.61 \\
		\textbf{Ours} (V) & \textbf{9,993} & \textbf{7} & \textbf{0.07} & \textbf{99.93} \\
		\bottomrule
	\end{tabular}
	\vspace{2mm}

	\caption{\rev{False positives of TSC-Net on challenging real-video subsets. Each subset contains the 2,500 MSR-VTT clips at the top or bottom quartile of a fixed statistic computed on the 8-frame input window; subsets overlap.}}
	\label{table:fpr_stress}
	\setlength{\tabcolsep}{4pt}
	\begin{tabular}{>{\raggedright\arraybackslash}p{3.7cm} >{\raggedright\arraybackslash}p{6.0cm} c c c}
		\toprule
		\textbf{Targeted stress case} & \textbf{Operational proxy} & \textbf{\#Videos} & \textbf{FP} & \textbf{FPR (\%)} \\
		\midrule
		Strong motion / temporal change & Top-25\% mean absolute grayscale inter-frame difference & 2,500 & 0 & 0.00 \\
		High contrast & Top-25\% RMS contrast & 2,500 & 1 & 0.04 \\
		Complex high-frequency background & Top-25\% mean Sobel magnitude & 2,500 & 5 & 0.20 \\
		Camera motion / shake & Top-25\% global-transform magnitude & 2,500 & 1 & 0.04 \\
		Natural local temporal irregularity & Top-25\% residual variation after global compensation & 2,500 & 0 & 0.00 \\
		Weak motion & Bottom-25\% mean absolute grayscale inter-frame difference & 2,500 & 3 & 0.12 \\
		\bottomrule
	\end{tabular}
	\vspace{-\intextsep}
\end{table}
\noindent\textbf{Protocol.}
Since false positives are critical for real-world deployment, we evaluate 10,000 real MSR-VTT web clips (7,180 source videos, 257 queries, 20 categories).
Taking generated videos as positive, $\mathrm{FPR}=\mathrm{FP}/(\mathrm{FP}+\mathrm{TN})$ and specificity $=\mathrm{TN}/(\mathrm{TN}+\mathrm{FP})$.
Each method retains its original score direction and uses a fixed threshold of $0.5$, without test-set tuning.
As shown in Table~\ref{table:fpr_full}, TSC-Net produces 7 false positives overall (FPR $0.07\%$, specificity $99.93\%$), the fewest among the evaluated methods.

\noindent\textbf{Challenging real-video categories.}
For each targeted stress case, we use a fixed statistic computed on the 8-frame input window as a proxy and form overlapping quartile subsets (Table~\ref{table:fpr_stress}).
Together, these statistics provide reproducible proxies for strong and weak motion, contrast, spatial complexity, camera-induced motion, and local non-rigid variation after the dominant camera motion is removed; they are not manually verified semantic categories.
TSC-Net produces at most 5 false positives (FPR $0.20\%$) in any subset, with the highest count on the complex high-frequency background subset.

\noindent\textbf{Natural high-frequency dynamics vs.\ generative flickering.}
Both real and generated videos can contain rich high-frequency content; what distinguishes them is how this content evolves over time.
As visualized in Fig.~\ref{fig:intro_abc}(a), a real video may contain a complex high-frequency background, yet its local patterns evolve continuously and form relatively coherent trajectories in the temporal slices.
In an AI-generated video, flickering textures, unstable boundaries, or abruptly changing details may interrupt this continuity, producing distorted, broken, or irregular slice structures.
In TSC-Net, the gradient magnitude only determines where slices are extracted and is never used as a classification score.
Raw slices preserve the temporal trajectories, first- and second-order difference slices make abrupt changes and unstable evolution more explicit, and the Transformer models dependencies across time, selected rows and columns, spatial directions, and HF/LF groups.
The model can therefore learn whether high-frequency details evolve coherently, rather than treating high-frequency energy itself as evidence of generation.
Consistent with this observation, Fig.~\ref{fig:intro_abc}(b) shows that real and generated distributions strongly overlap in the patch domain (AUROC $0.502$/$0.525$ for HF/LF patches), whereas reorganizing the same content into temporal slices increases the AUROC to $0.770$/$0.650$.

\noindent\textbf{Weak-motion videos.}
To examine naturally weak motion, we separately rank the MSR-VTT real videos and the ten GenVideo generated-video pools by the mean absolute grayscale inter-frame difference, and select the 2,500 lowest-variation videos from each class.
On this balanced weak-motion set, TSC-Net obtains 0.9528/0.9969/0.9960 ACC/AP/AUROC, compared with 0.9552/0.9957/0.9953 for ReStraV.
On this weak-motion subset, TSC-Net remains comparable to ReStraV. A visually static generated scene may still contain local background or detail deformation (Fig.~\ref{fig:intro_abc}(a)).
Videos with almost no temporal variation nevertheless remain a limitation of our method, as discussed in Sec.~\ref{app:limitations}.
\end{revblock}
\begin{figure}[!t]
	\centering
	\footnotesize
	\includegraphics[width=0.6\linewidth]{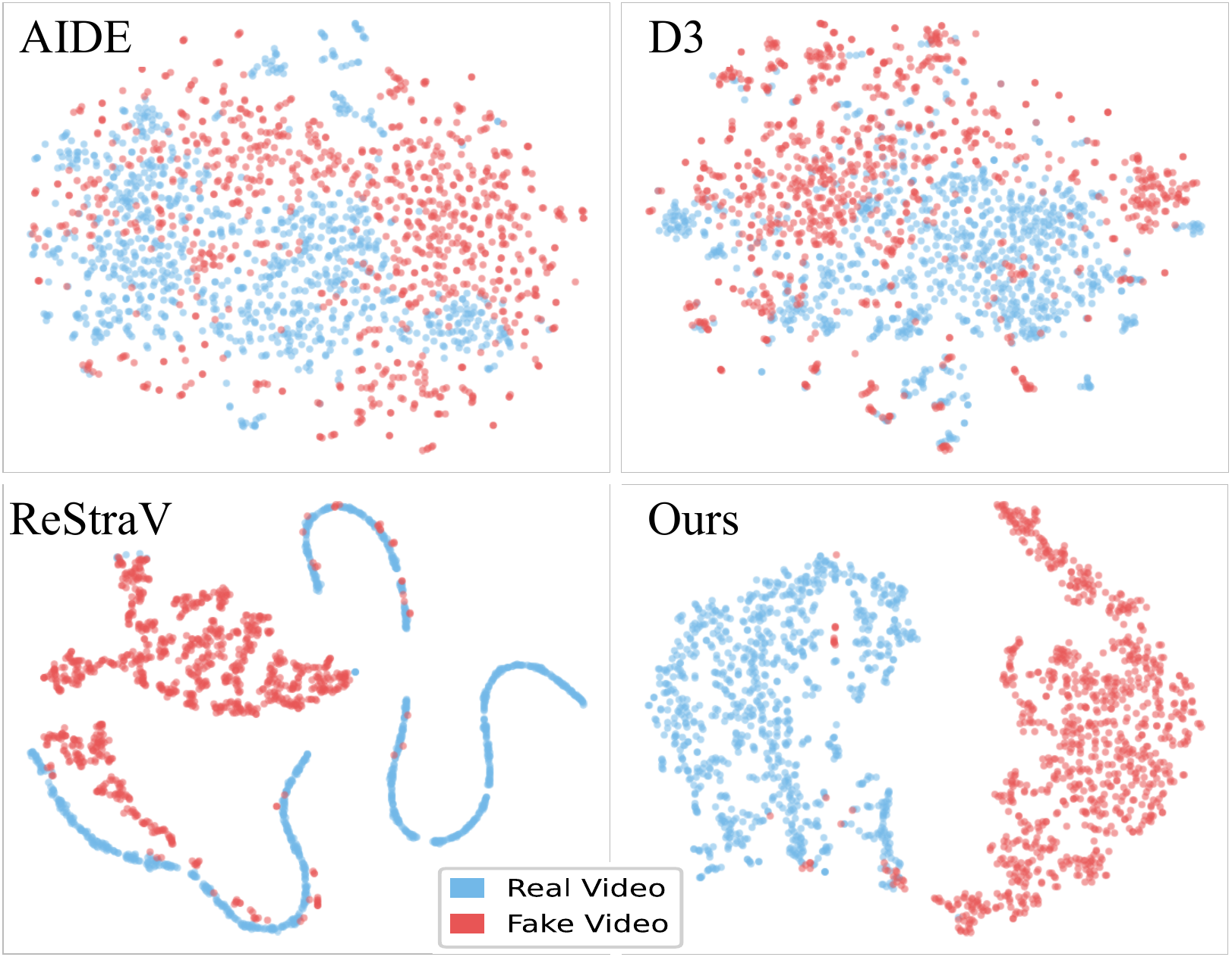}
	\caption{t-SNE visualization of features extracted from GenVideo using representative state-of-the-art methods and our method.}
	\label{fig:tsne}
	\vspace{-\intextsep}
\end{figure}

\section{Feature Space Visualization}
\label{app:feature_visualization}

To qualitatively analyze the learned representations, we visualize the feature space using t-SNE~\cite{van2008visualizing} in Fig.~\ref{fig:tsne}. 
We randomly sample 100 videos from each generation source, resulting in 1000 samples in total, and compare our method with several strong baselines, including AIDE, D3, and ReStraV. 
Compared with prior methods, our method shows more compact intra-class clustering and clearer inter-class separation in the projected space. 
Although t-SNE is only a qualitative visualization tool, this trend is consistent with the quantitative improvements reported in the main manuscript, suggesting that the proposed temporal slice representation provides more discriminative features for distinguishing real and AI-generated videos.

\section{Limitations and Future Work}
\label{app:limitations}
\begin{figure}[!t]
	\centering
	\footnotesize
	\includegraphics[width=0.8\linewidth]{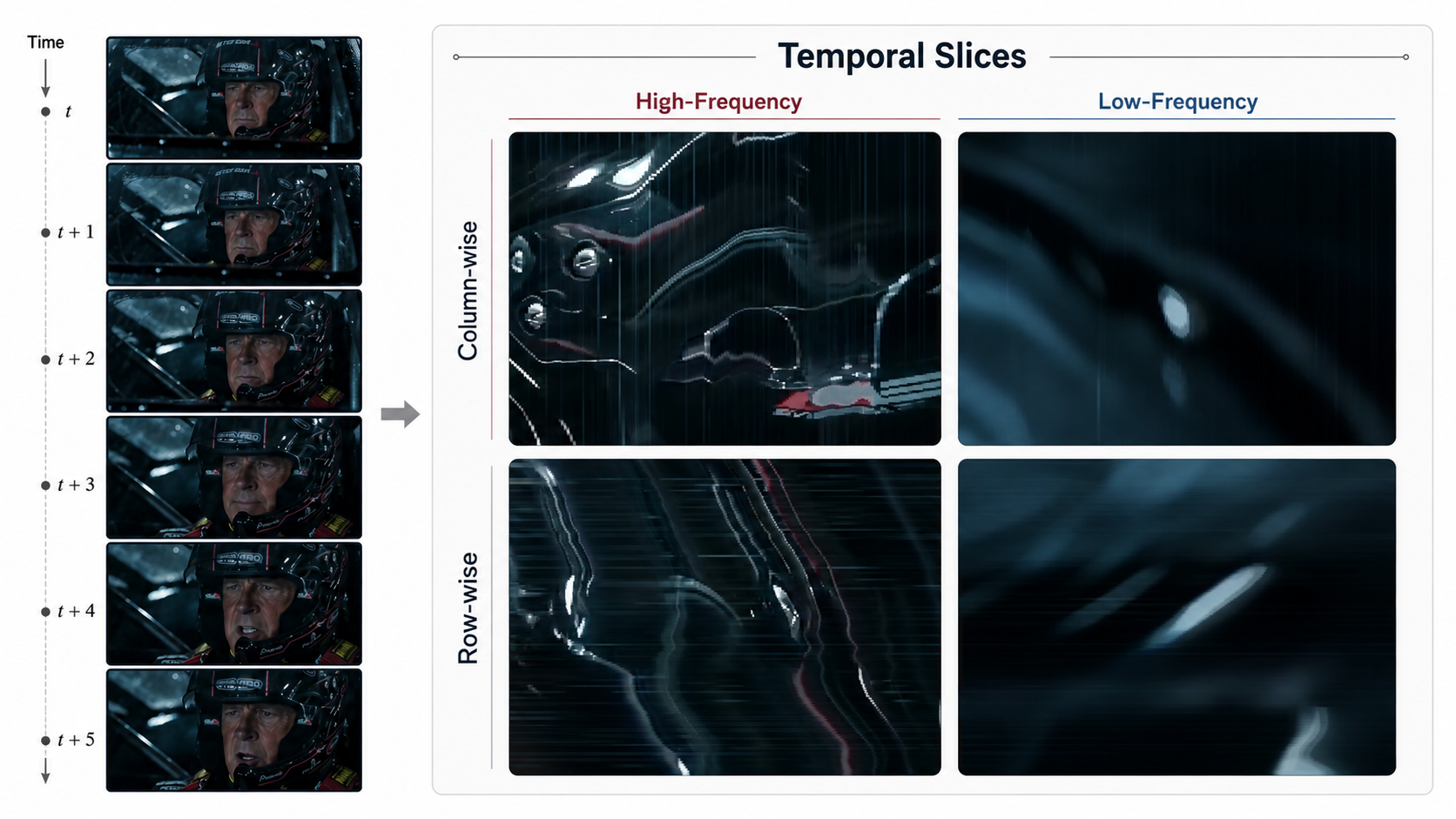}
	\caption{Failure case on a highly realistic video in Seedance (Film). 
		Left: sampled frames from the input video. 
		Right: row-wise and column-wise temporal slices extracted from high-frequency and low-frequency regions. 
		Although the slice representations preserve temporal evolution, the video \rev{shows only} subtle facial expression changes with limited temporal variation, making both high-frequency and low-frequency slices weakly discriminative. 
		As a result, all compared methods, including ours, fail on this example.}
	\label{fig:failure_case}
	\vspace{-\intextsep}
\end{figure}

Despite its effectiveness, our method still has several limitations. 
First, our framework relies on temporal slices to expose temporal inconsistencies. 
As shown in Fig.~\ref{fig:failure_case}, this assumption becomes less effective for highly realistic videos in Seedance (Film), where \rev{facial expression changes are subtle} and the temporal evolution remains very weak. 
In such cases, both high-frequency and low-frequency temporal slices may contain limited discriminative cues, making the video difficult to distinguish even for our method. 
More broadly, this example also reflects the increasing challenge posed by recent high-fidelity generators, for which even existing state-of-the-art detectors may fail.
\rev{On a balanced set of 5,000 low-variation videos, our method remains comparable to ReStraV (Sec.~\ref{app:fpr}).}
Second, the current slice construction strategy is still based on hand-crafted heuristics (e.g., gradient-based patch selection), which may not always be optimal for diverse video content. 
\rev{A semantically guided Grad-CAM localizer did not improve this accuracy--efficiency trade-off in our experiments (Sec.~\ref{app:ablation_loc}).}
Finally, our framework primarily focuses on spatial--temporal consistency and does not explicitly leverage other potentially useful cues, such as semantic priors, audio signals, or metadata.

In future work, we plan to explore more adaptive and learnable slice selection mechanisms, and to combine slice-based temporal modeling with complementary cues beyond temporal inconsistency alone. 
This may include integrating semantic representations, multimodal signals, or other auxiliary information to better handle videos with only weak temporal variation. 
We also aim to continuously expand the AI-Artist benchmark to track the rapid evolution of generative models and provide a more comprehensive evaluation platform for the community.

\rev{For weak-motion videos, we could explore training-time knowledge distillation from a full-video teacher, after verifying that the teacher performs better on this subset. Its frame representations and video-level predictions could complement the local slice cues. To account for differences between full-video and slice representations, distillation could match temporal similarities and relative changes between frame representations rather than feature vectors, with reduced supervision when adjacent teacher representations change too little. Removing the teacher and auxiliary modules after training would leave TSC-Net's inference architecture, parameter count, and computational cost unchanged. If audio is available, a shallow encoder could extract log-Mel features from the same temporal window, align them with the slice features, and fuse them through a lightweight gate. Because authentic videos may be edited, dubbed, or desynchronized, audio-visual mismatch should complement the visual evidence rather than serve as an independent cue for generation. Videos without audio would use the original visual model.}

\section{Potential Societal Impacts}
\label{app:impacts}
The ability to reliably detect AI-generated videos has important societal implications. 
On the positive side, our work contributes to improving media authenticity and mitigating the risks associated with the misuse of generative models, such as misinformation, fraud, and malicious content creation. 
By providing a more efficient and accurate detection framework, our method can facilitate large-scale deployment in real-world scenarios, including content moderation, digital forensics, and platform-level integrity protection. 
In addition, the introduction of the \textit{AI-Artist} dataset, which contains videos generated by recent high-fidelity models, helps the community better evaluate detection systems under more realistic and challenging conditions, promoting more robust and future-proof solutions.

However, our work also raises several potential concerns. 
First, like many detection approaches, our method may be used in an adversarial setting, where advances in detection could incentivize the development of more sophisticated generation techniques, potentially leading to an arms race between generation and detection. 
Second, although the AI-Artist dataset is constructed for research purposes, it may still contain synthetic content that resembles real individuals or scenes, which could raise concerns about privacy, consent, or misuse if redistributed without proper safeguards. 
Third, automated detection systems are not infallible and may produce false positives or false negatives, which could have downstream consequences in high-stakes applications such as content moderation or legal decision-making.

To mitigate these risks, we emphasize that our dataset is curated with a focus on synthetic content and is intended strictly for research use. 
We encourage responsible deployment of detection systems with human oversight, transparency, and appropriate evaluation across diverse scenarios. 
Future work should also consider fairness, robustness, and accountability, ensuring that detection technologies are developed and used in a manner that aligns with societal values.

\section{Visualization of AI-Artist}
\label{app:visualization_data}
\begin{figure}[!hbt]\footnotesize
	\centering
	\includegraphics[width=\linewidth]{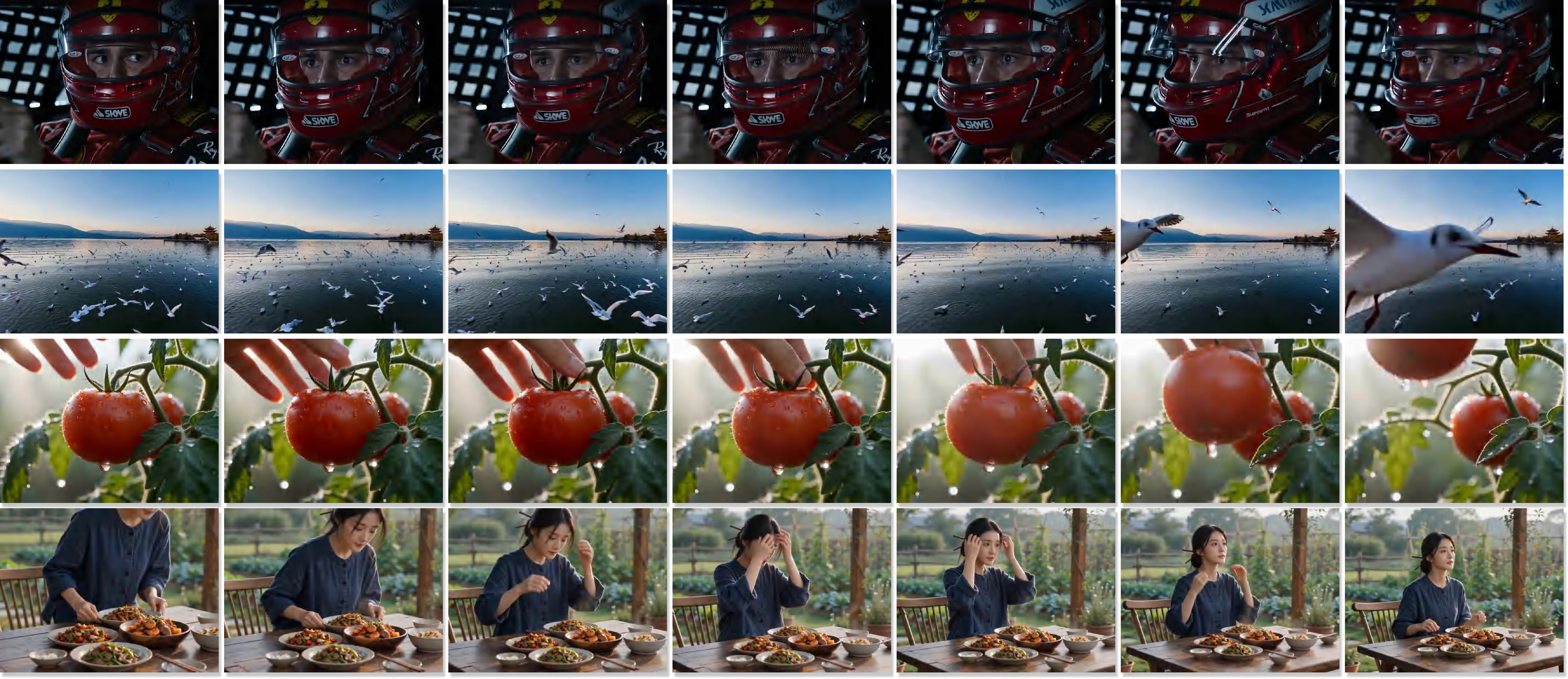} 
	\caption{Visualization of Seedance (Film) videos collected from public online communities.}
	\label{fig:sample1}
	\vspace{-\intextsep}
\end{figure}

\begin{figure}[!hbt]\footnotesize
	\centering
	\includegraphics[width=\linewidth]{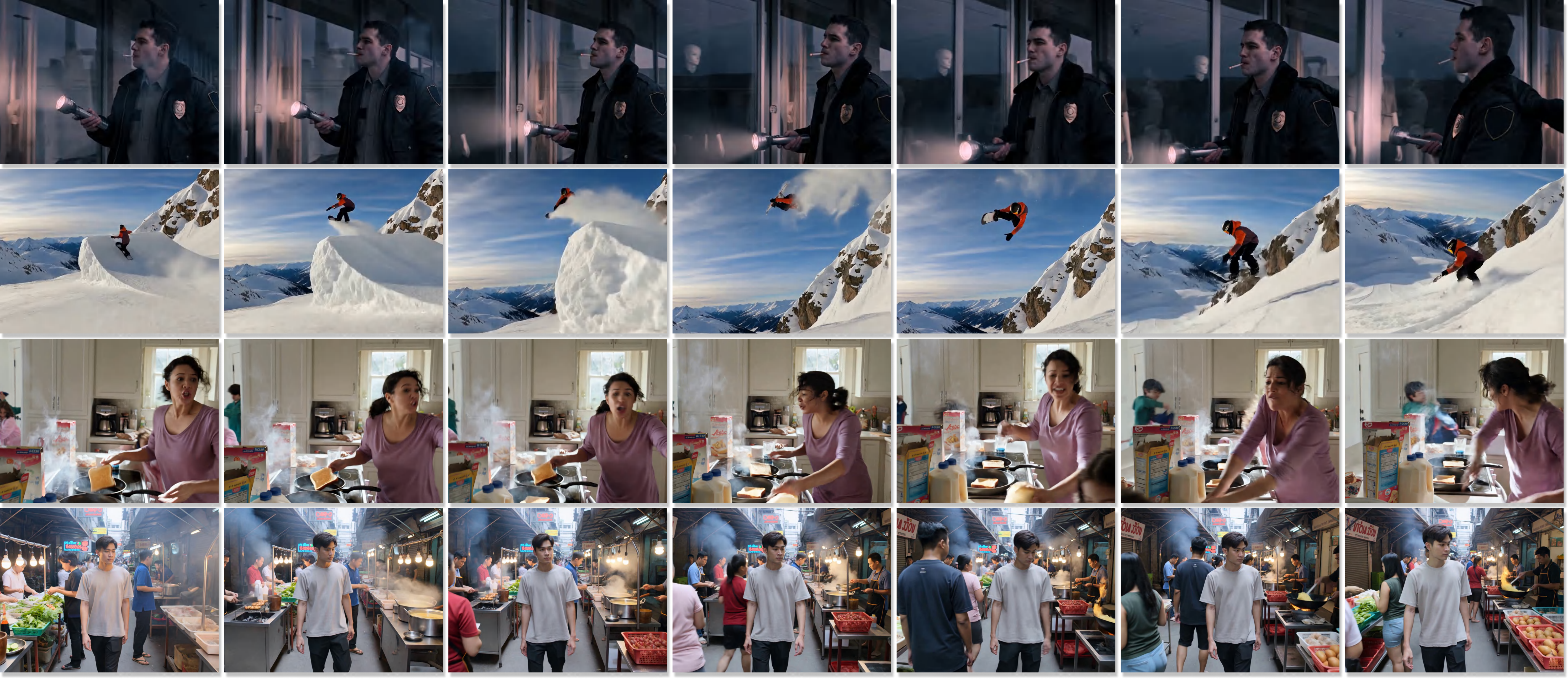} 
	\caption{Visualization of Kling 3.0 videos collected from official showcases and public creator communities.}
	\label{fig:sample2}
	\vspace{-\intextsep}
\end{figure}

\begin{figure}[!hbt]\footnotesize
	\centering
	\includegraphics[width=\linewidth]{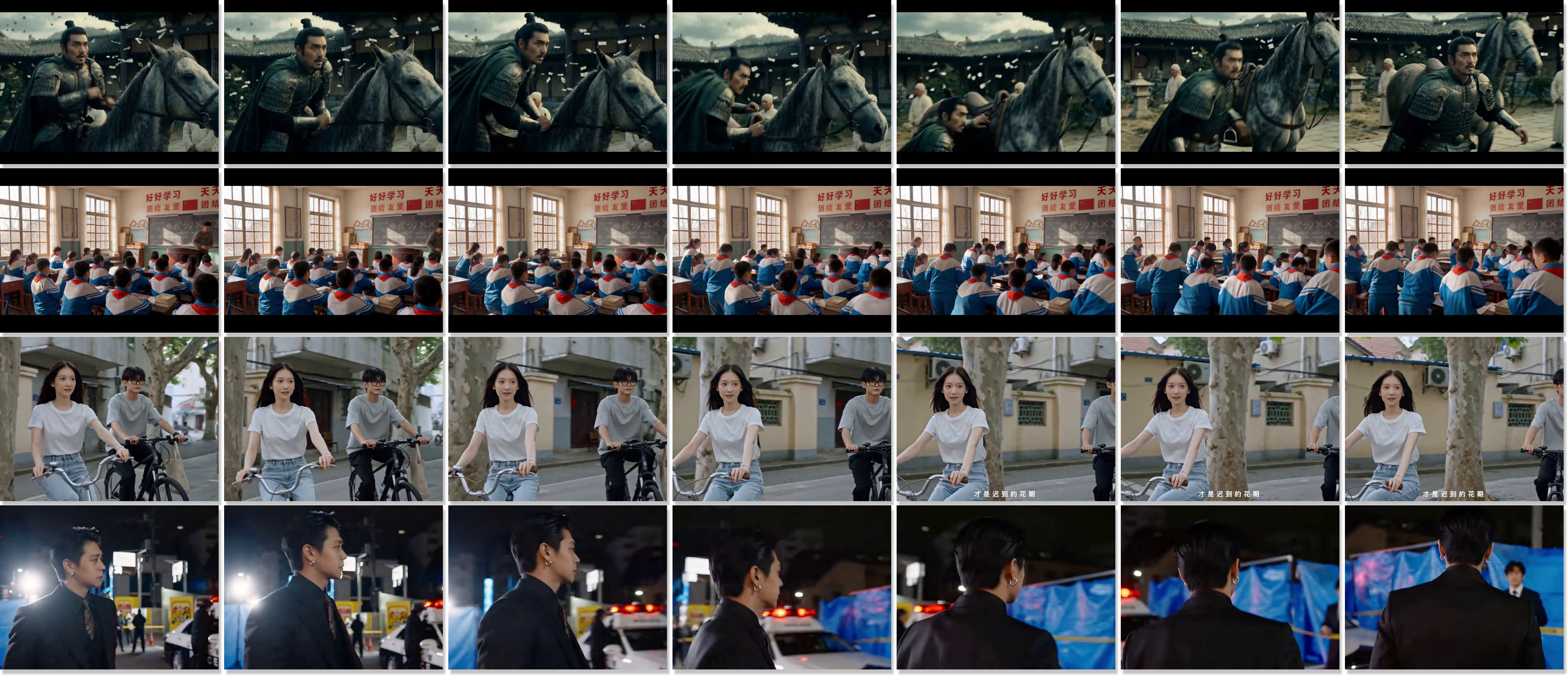} 
	\caption{Visualization of AI-generated creator videos collected from Douyin public platforms.}
	\label{fig:sample3}
	\vspace{-\intextsep}
\end{figure}

\begin{figure}[!hbt]\footnotesize
	\centering
	\includegraphics[width=\linewidth]{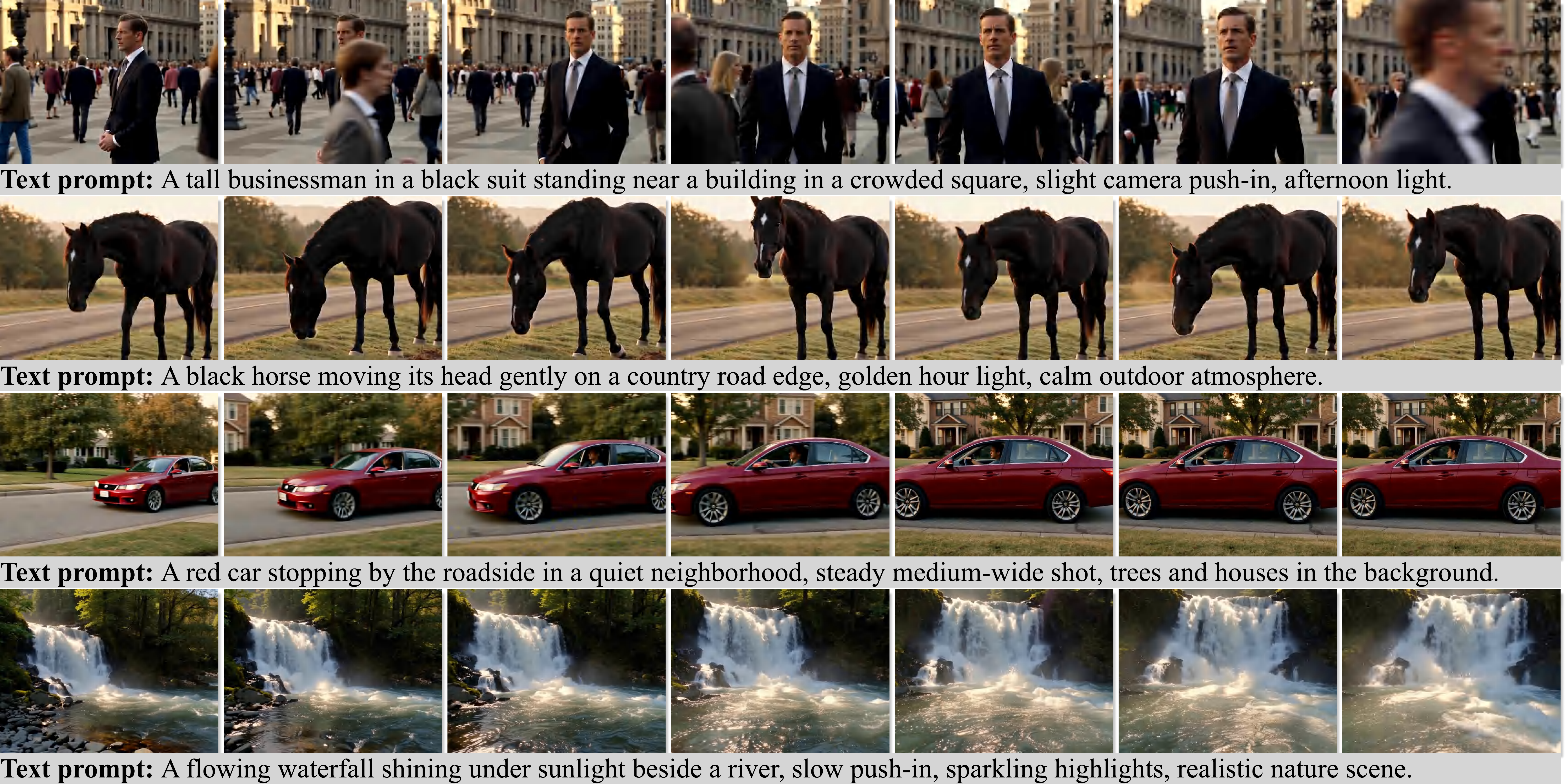} 
	\caption{Visualization of Runway Gen-4.5~\cite{Runway2025} generated samples.}
	\label{fig:sample4}
	\vspace{-\intextsep}
\end{figure}

\begin{figure}[!hbt]\footnotesize
	\centering
	\includegraphics[width=\linewidth]{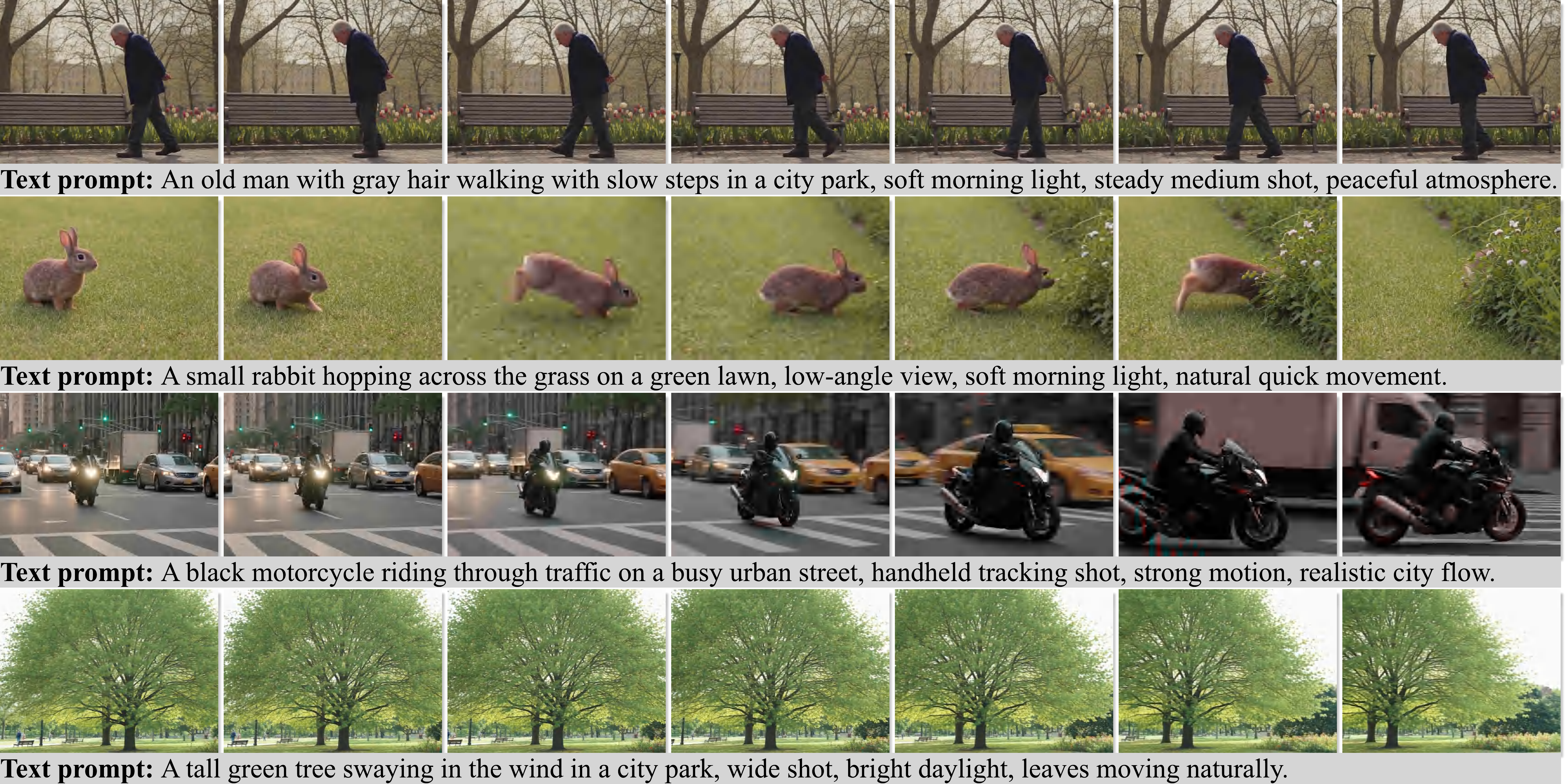} 
	\caption{Visualization of PixVerse V5.6~\cite{PixVerse2026} generated samples.}
	\label{fig:sample5}
	\vspace{-\intextsep}
\end{figure}

\begin{figure}[!hbt]\footnotesize
	\centering
	\includegraphics[width=\linewidth]{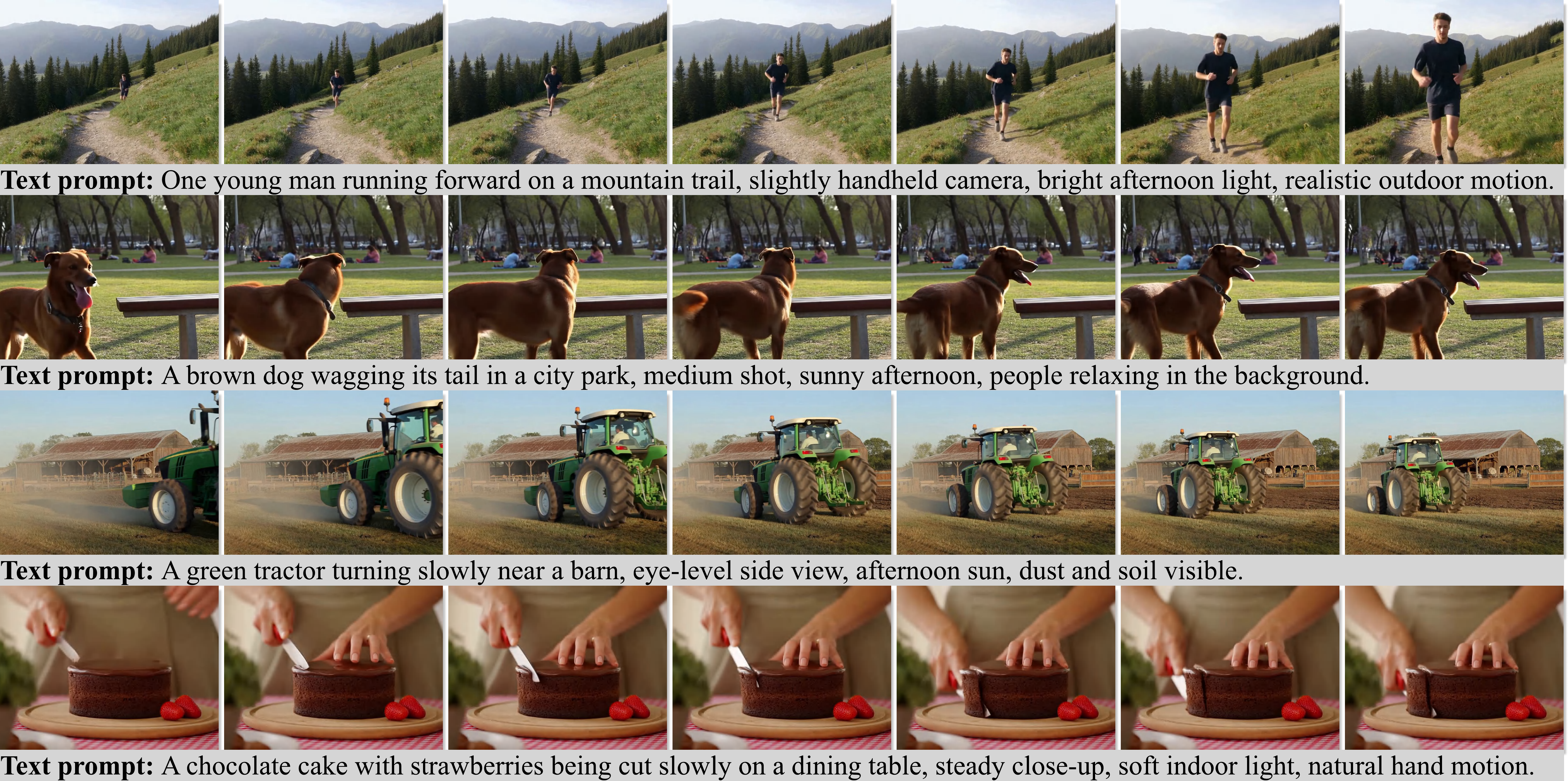} 
	\caption{Visualization of Hailuo 2.3~\cite{Hailuo2025} generated samples.}
	\label{fig:sample6}
	\vspace{-\intextsep}
\end{figure}

\begin{figure}[!hbt]\footnotesize
	\centering
	\includegraphics[width=\linewidth]{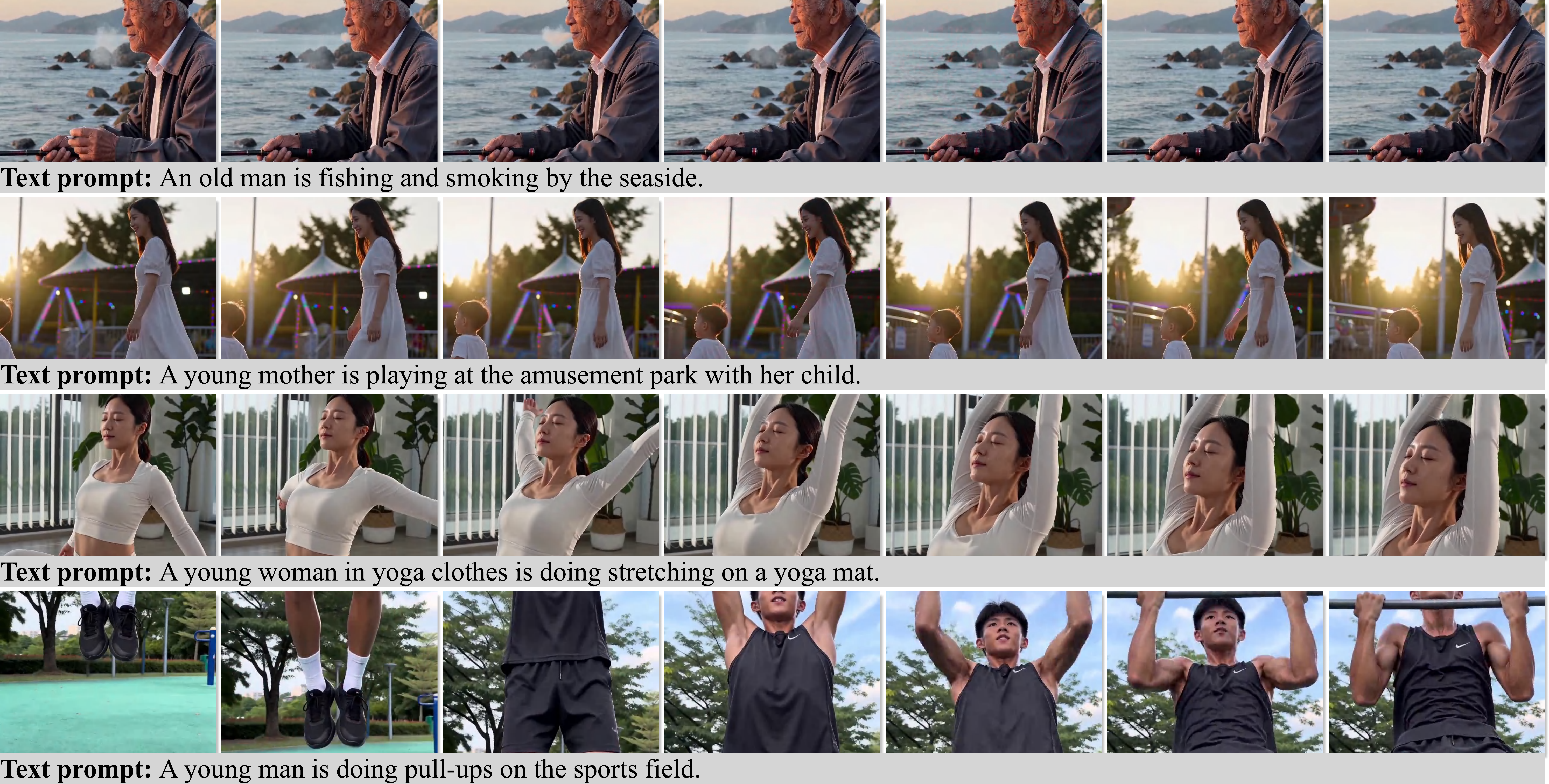} 
	\caption{Visualization of Seedance 2.0~\cite{seedance2026seedance} generated samples.}
	\label{fig:sample7}
	\vspace{-\intextsep}
\end{figure}

\begin{figure}[!hbt]\footnotesize
	\centering
	\includegraphics[width=\linewidth]{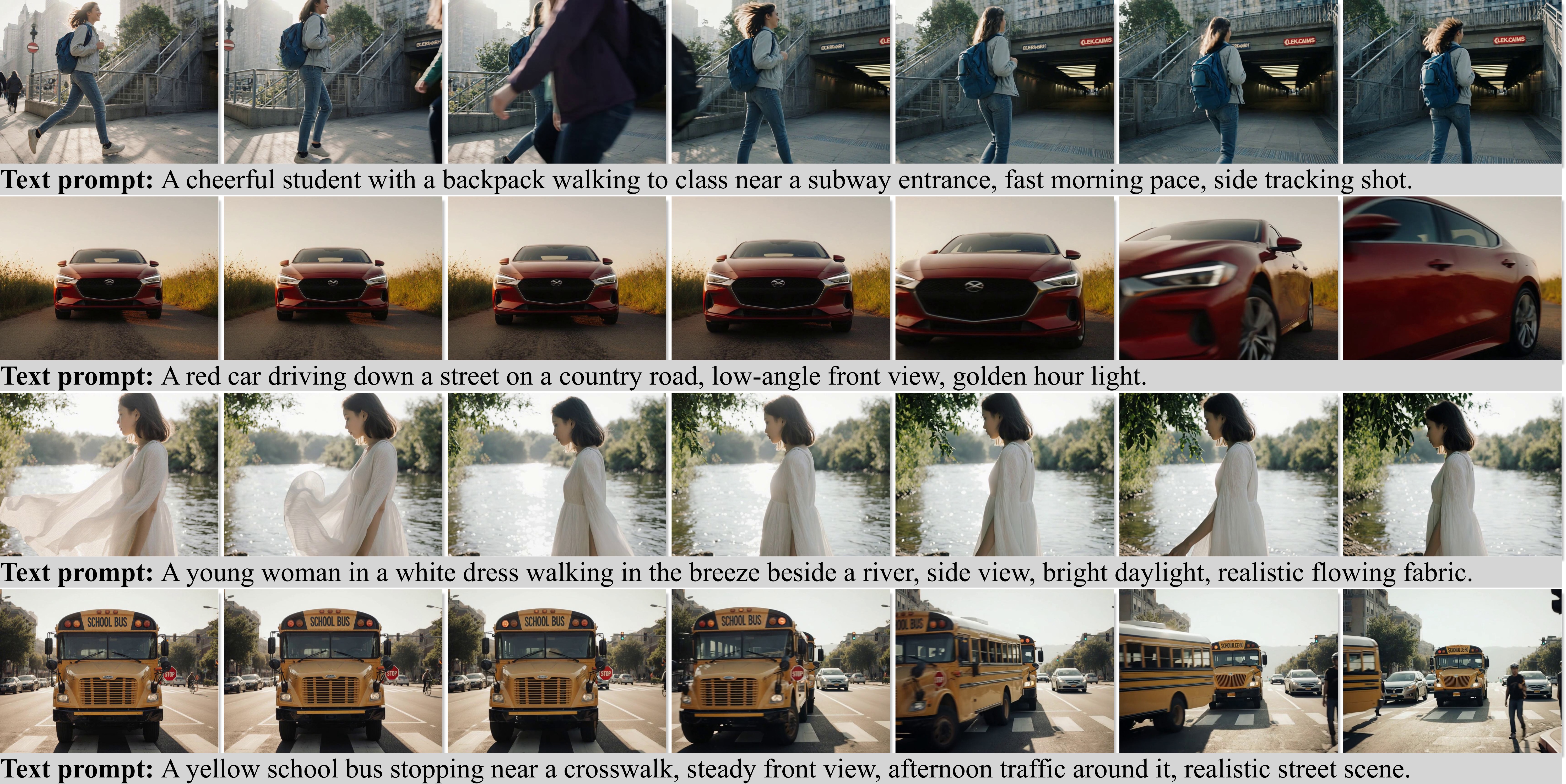} 
	\caption{Visualization of Luma Ray3.14~\cite{Luma2026} generated samples.}
	\label{fig:sample8}
	\vspace{-\intextsep}
\end{figure}

Following the collection details in Sec.~\ref{app:collection_data}, we further provide qualitative examples to illustrate the visual characteristics of AI-Artist. As shown in Figs.~\ref{fig:sample1}--~\ref{fig:sample8}, the dataset contains diverse samples from both the \textit{web-collected} and \textit{self-generated} branches, covering a wide range of subjects, actions, scenes, camera motions, durations, resolutions, and visual styles. More video examples are provided in the supplementary material.

\clearpage
\newpage
\section*{NeurIPS Paper Checklist}

\begin{enumerate}

\item {\bf Claims}
    \item[] Question: Do the main claims made in the abstract and introduction accurately reflect the paper's contributions and scope?
    \item[] Answer: \answerYes{} %
    \item[] Justification: 
    The main claims in the abstract and introduction are aligned with the scope of the paper. They are supported by the proposed temporal-slice representation, the TSC-Net architecture, and the empirical results on GenVideo and AI-Artist, including both detection performance and computational efficiency.

\item {\bf Limitations}
    \item[] Question: Does the paper discuss the limitations of the work performed by the authors?
    \item[] Answer: \answerYes{} %
    \item[] Justification: The paper includes a dedicated limitations section in the appendix. It discusses the scope of the proposed method, potential dataset and deployment limitations, and the need for further evaluation under future generative models and real-world perturbations.

\item {\bf Theory assumptions and proofs}
    \item[] Question: For each theoretical result, does the paper provide the full set of assumptions and a complete (and correct) proof?
    \item[] Answer: \answerNA{}  %
    \item[] Justification: The paper does not make formal theoretical claims or provide theorem-proof results. Our contribution is an insight-driven representation design: we reformulate AI-generated video detection using compact frequency-aware temporal slices, and support this formulation with analysis and empirical validation.

    \item {\bf Experimental result reproducibility}
    \item[] Question: Does the paper fully disclose all the information needed to reproduce the main experimental results of the paper to the extent that it affects the main claims and/or conclusions of the paper (regardless of whether the code and data are provided or not)?
    \item[] Answer: \answerYes{}  %
    \item[] Justification: The paper provides the details necessary to reproduce the main experimental results, including the training and test datasets, evaluation metrics, preprocessing pipeline, model hyperparameters, optimizer, training schedule, and implementation platform. Additional dataset construction details and ablation studies are provided in the appendix.

\item {\bf Open access to data and code}
    \item[] Question: Does the paper provide open access to the data and code, with sufficient instructions to faithfully reproduce the main experimental results, as described in supplemental material?
    \item[] Answer: \rev{\answerNo{}}
    \item[] Justification: The source code \rev{was developed within an industry-collaboration project and is subject to the corresponding intellectual-property and confidentiality agreements; therefore, it cannot} be publicly released. \rev{To support reproducibility, the paper provides detailed descriptions of the preprocessing pipeline, temporal slice construction, network architecture, hyperparameters, training protocol, and evaluation settings in Secs.~2--3 and the appendix. The GenVideo training and test data are publicly available, and the construction protocol and statistics of AI-Artist are documented in the appendix.} \rev{The AI-Artist benchmark will be publicly released for research purposes through the project page, subject to the usage terms of the source platforms and generation services.}

\item {\bf Experimental setting/details}
    \item[] Question: Does the paper specify all the training and test details (e.g., data splits, hyperparameters, how they were chosen, type of optimizer) necessary to understand the results?
    \item[] Answer: \answerYes{}  %
    \item[] Justification: The paper specifies the training and test settings, including the GenVideo training protocol, the GenVideo OOD test set, the newly constructed AI-Artist dataset, evaluation metrics, preprocessing details, model hyperparameters, optimizer, learning rate schedule, batch size, number of training iterations, and hardware platform.

\item {\bf Experiment statistical significance}
    \item[] Question: Does the paper report error bars suitably and correctly defined or other appropriate information about the statistical significance of the experiments?
    \item[] Answer: \answerNo{}  %
    \item[] Justification: The paper reports extensive comparisons across multiple datasets, generators, perturbation settings, and ablation variants. Therefore, including error bars is not strictly necessary.

\item {\bf Experiments compute resources}
    \item[] Question: For each experiment, does the paper provide sufficient information on the computer resources (type of compute workers, memory, time of execution) needed to reproduce the experiments?
    \item[] Answer: \answerYes{}  %
    \item[] Justification: The paper reports the hardware platform used for the experiments and provides model efficiency in terms of parameters, GFLOPs, and average inference time per video. The main experiments were conducted on an NVIDIA RTX PRO 6000 GPU using PyTorch.
    
\item {\bf Code of ethics}
    \item[] Question: Does the research conducted in the paper conform, in every respect, with the NeurIPS Code of Ethics \url{https://neurips.cc/public/EthicsGuidelines}?
    \item[] Answer: \answerYes{}  %
    \item[] Justification: The research conforms to the NeurIPS Code of Ethics. The work focuses on detecting AI-generated videos for media authenticity and forensic analysis, and we discuss potential societal risks and responsible-use considerations in the appendix.

\item {\bf Broader impacts}
    \item[] Question: Does the paper discuss both potential positive societal impacts and negative societal impacts of the work performed?
    \item[] Answer: \answerYes{}  %
    \item[] Justification: The paper discusses both positive and negative societal impacts. It highlights potential benefits for media authenticity, content moderation, and digital forensics, while also discussing risks such as adversarial misuse, privacy concerns, and possible false positives or false negatives in high-stakes applications.
    
\item {\bf Safeguards}
    \item[] Question: Does the paper describe safeguards that have been put in place for responsible release of data or models that have a high risk for misuse (e.g., pre-trained language models, image generators, or scraped datasets)?
    \item[] Answer: \answerYes{}  %
    \item[] Justification: The paper discusses safeguards for responsible use of the proposed dataset and detector. The AI-Artist dataset is curated for research purposes, and we emphasize responsible deployment with human oversight, transparency, and evaluation across diverse scenarios.

\item {\bf Licenses for existing assets}
    \item[] Question: Are the creators or original owners of assets (e.g., code, data, models), used in the paper, properly credited and are the license and terms of use explicitly mentioned and properly respected?
    \item[] Answer: \answerYes{}  %
    \item[] Justification: The paper cites the existing datasets, models, and baselines used in the experiments, including GenVideo, Youku-mPLUG, MSR-VTT, and representative detection baselines. For web-collected or commercial-model-generated videos in AI-Artist, we describe the collection protocol and use them for research evaluation under the corresponding platform and model usage constraints.

\item {\bf New assets}
    \item[] Question: Are new assets introduced in the paper well documented and is the documentation provided alongside the assets?
    \item[] Answer: \answerYes{}  %
    \item[] Justification: The paper introduces AI-Artist as a new benchmark and documents its construction protocol, including the web-collected and self-generated subsets, source types, generation models, prompt construction strategy, semantic coverage, and dataset statistics. Qualitative examples and additional details are provided in the appendix. \rev{The benchmark and its documentation will be made available through the project page.}

\item {\bf Crowdsourcing and research with human subjects}
    \item[] Question: For crowdsourcing experiments and research with human subjects, does the paper include the full text of instructions given to participants and screenshots, if applicable, as well as details about compensation (if any)? 
    \item[] Answer: \answerNA{}  %
    \item[] Justification: The paper does not involve crowdsourcing experiments or research with human subjects. The dataset consists of AI-generated videos collected from public online sources or generated using commercial video generation models.

\item {\bf Institutional review board (IRB) approvals or equivalent for research with human subjects}
    \item[] Question: Does the paper describe potential risks incurred by study participants, whether such risks were disclosed to the subjects, and whether Institutional Review Board (IRB) approvals (or an equivalent approval/review based on the requirements of your country or institution) were obtained?
    \item[] Answer: \answerNA{}  %
    \item[] Justification: The paper does not involve human-subject experiments, user studies, or crowdsourcing. Therefore, IRB approval or an equivalent review is not applicable.

\item {\bf Declaration of LLM usage}
    \item[] Question: Does the paper describe the usage of LLMs if it is an important, original, or non-standard component of the core methods in this research? Note that if the LLM is used only for writing, editing, or formatting purposes and does \emph{not} impact the core methodology, scientific rigor, or originality of the research, declaration is not required.
    \item[] Answer: \answerYes{}  %
    \item[] Justification: The paper uses a large language model only for constructing and expanding prompts in the self-generated subset of AI-Artist. The LLM is not part of the proposed detector, model architecture, training objective, inference pipeline, or any component that contributes to the core methodological novelty.

\end{enumerate}

\end{document}